\documentclass[10pt,journal,compsoc]{IEEEtran}

\usepackage{booktabs} 

\usepackage{xcolor}
\usepackage{ifthen}
\usepackage{microtype}
\PassOptionsToPackage{warn}{textcomp}
\usepackage{textcomp} 
\usepackage{url}

\usepackage{cite}
\usepackage{tabu}
\usepackage{booktabs}
\usepackage{amsmath}
\usepackage{amssymb}
\usepackage{subfigure}
\usepackage{graphicx}
\usepackage{soul}
\usepackage{algorithmic}
\usepackage{algorithm}
\usepackage{ragged2e}

\usepackage{bbding}

\usepackage{color}
\definecolor{quill}{rgb}{0.69,0.61,0.85}
\definecolor{contentchange}{rgb}{0.9,0.61,0.5}

\begin{document}

\title{CreativeSynth: Cross-Art-Attention for Artistic Image Synthesis with Multimodal Diffusion}

\author{Nisha Huang,
        Weiming Dong,~\IEEEmembership{Member,~IEEE},
        Yuxin Zhang,~\IEEEmembership{Graduate Student Member, ~IEEE},
        Fan Tang,
        Ronghui Li,
        Chongyang Ma,
        Xiu Li,~\IEEEmembership{Member,~IEEE},
        Tong-Yee Lee,~\IEEEmembership{Senior Member,~IEEE},
        Changsheng Xu,~\IEEEmembership{Fellow,~IEEE}
\thanks{This work was supported in part by the Beijing Science and Technology Plan Project under no. Z231100005923033, in part by the Beijing Natural Science Foundation under Grant L221013, and in part by the National Science and Technology Council under no. 113-2221-E-006-161-MY3, Taiwan.
(Corresponding author: Weiming Dong.)}
\IEEEcompsocitemizethanks{
\IEEEcompsocthanksitem N. Huang is with Tsinghua Shenzhen International Graduate School, Tsinghua University, Shenzhen 518071, China, and also with Pengcheng Laboratory, Shenzhen 518055, China. E-mail: hns24@mails.tsinghua.edu.cn.
\IEEEcompsocthanksitem N. Huang, W. Dong, Y. Zhang and C. Xu are with MAIS, Institute of Automation, Chinese Academy of Sciences, Beijing 100190, China, and also with School of Artificial Intelligence, University of Chinese Academy of Sciences, Beijing 100049, China. E-mail:\{huangnisha2021,zhangyuxin2020,weiming.dong\}@ia.ac.cn and csxu@nlpr.ia.ac.cn.
\IEEEcompsocthanksitem F. Tang is with the University of Chinese Academy of Sciences, Beijing 100049, China. E-mail: tfan.108@gmail.com.
\IEEEcompsocthanksitem R. Li and X. Li are with Tsinghua Shenzhen International Graduate School, Tsinghua University, Shenzhen 518071, China. E-mail:  lrh22@mails.tsinghua.edu.cn and li.xiu@sz.tsinghua.edu.cn.
\IEEEcompsocthanksitem C. Ma is with ByteDance Inc. E-mail: chongyangm@gmail.com.
\IEEEcompsocthanksitem T.-Y. Lee is with National Cheng Kung University, Tainan 701, Taiwan.\protect\\
E-mail: tonylee@mail.ncku.edu.tw.}
}

\markboth{IEEE TRANSACTIONS ON VISUALIZATION AND COMPUTER GRAPHICS~2025}%
{Huang \MakeLowercase{\textit{et al.}}:CreativeSynth: Cross-Art-Attention for Artistic Image Synthesis with Multimodal Diffusion}

\IEEEtitleabstractindextext{
\begin{abstract}
\justifying
Although remarkable progress has been made in image style transfer, style is just one of the components of artistic paintings. Directly transferring extracted style features to natural images often results in outputs with obvious synthetic traces. This is because key painting attributes including layout, perspective, shape, and semantics often cannot be conveyed and expressed through style transfer. Large-scale pretrained text-to-image generation models have demonstrated their capability to synthesize a vast amount of high-quality images. However, even with extensive textual descriptions, it is challenging to fully express the unique visual properties and details of paintings. Moreover, generic models often disrupt the overall artistic effect when modifying specific areas, making it more complicated to achieve a unified aesthetic in artworks. Our main novel idea is to integrate multimodal semantic information as a synthesis guide into artworks, rather than transferring style to the real world. We also aim to reduce the disruption to the harmony of artworks while simplifying the guidance conditions. Specifically, we propose an innovative multi-task unified framework called CreativeSynth, based on the diffusion model with the ability to coordinate multimodal inputs. CreativeSynth combines multimodal features with customized attention mechanisms to seamlessly integrate real-world semantic content into the art domain through Cross-Art-Attention for aesthetic maintenance and semantic fusion. We demonstrate the results of our method across a wide range of different art categories, proving that CreativeSynth bridges the gap between generative models and artistic expression. Code and results are available at https://github.com/haha-lisa/CreativeSynth.
\end{abstract}

\begin{IEEEkeywords}
Visual art, diffusion models, multimodal guidance, image generation.
\end{IEEEkeywords}}

\maketitle

\section{Introduction}
\label{sec:intro}
\textcolor{black}{
\IEEEPARstart{I}f a picture is worth a thousand words, then two pictures can weave a narrative beyond measure. In the field of artificial intelligence, a transformative era has dawned with the advent of text-to-image generation models~\cite{dreambooth,Key-Locked_SIGGRAPH23,gal2023encoder_TOG23,TI,dpl}, capable of creating vivid and contextually relevant visual representations from textual descriptions. These models exemplify the harmonious union of natural language understanding and the creation of art, fundamentally changing how we envision and materialize digital images, resonating with human creativity and intent. The inception of text-to-image models stems from a simple yet profound pursuit to transform words into the essence of images.}

\textcolor{black}{
Diffusion models~\cite{latentdiffusion,dalle2,glide} have become exemplary for text-guided image generation tasks, elegantly transforming latent noise into tangible, high-resolution visual content. However, directly applying these models for synthesizing and editing specific artistic images remains a challenge. Firstly, artistic images, lacking textual descriptions, contain subtle aspects such as specific styles, aesthetic significance, texture details, lighting conditions, and compositional perspectives~\cite{tvcg_EscherArts,tvcg_lineart,tvcg_VeCHArt,tvcg_animediffusion,ma2024taming}, which may not have corresponding terms in vocabularies. Secondly, even with textual prompts provided, existing text-to-image models tend to synthesize entirely new content without understanding the original work's creative direction, failing to follow the theme, composition, lighting, and style of the artistic image. In practical applications, especially those requiring complex control over visual properties and semantics, the application of such models is often hindered due to the limited influence of textual input on images. Lastly, users might wish to edit various images in numerous ways, thus tuning large models for every image and edit type due to high costs is undesirable.}

\begin{figure*}
\centering
\includegraphics[width=0.9\linewidth]{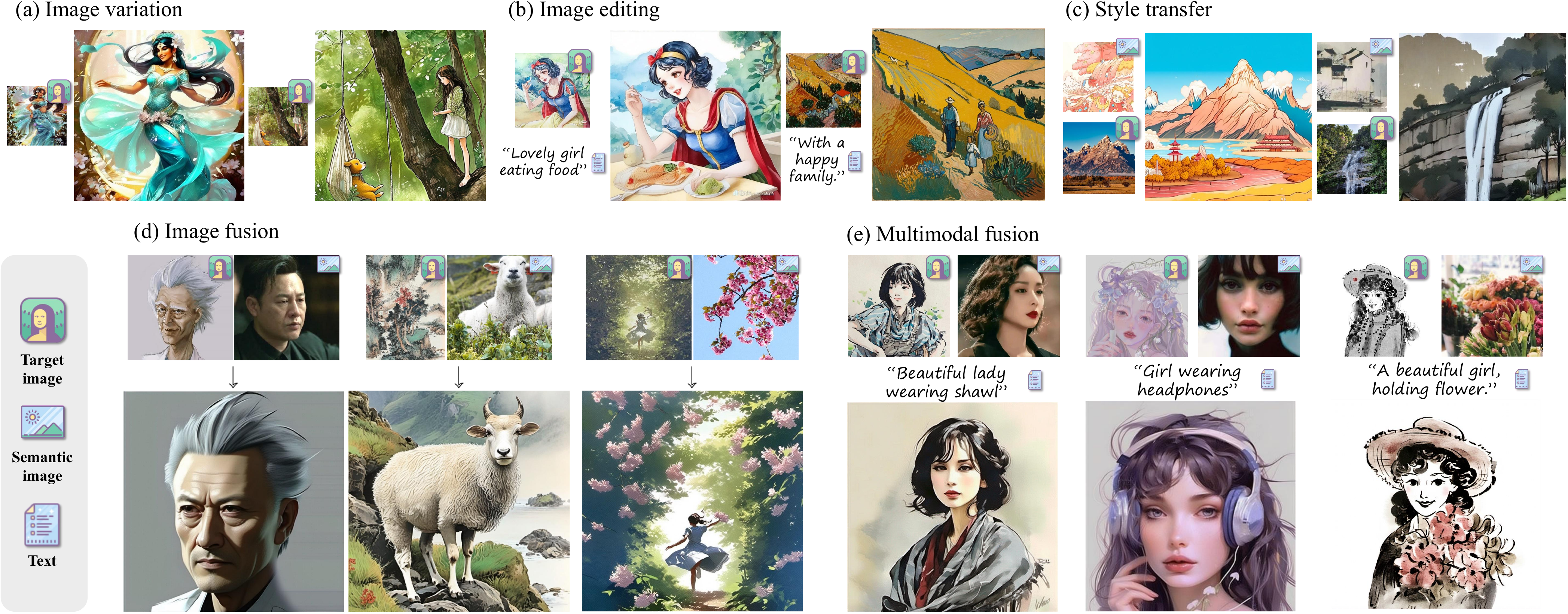}
\caption{\textcolor{black}{Our CreativeSynth unified framework is capable of generating personalized digital art when supplied with an art image, drawing on prompts from either unimodal or multimodal prompts. This methodology not only yields artwork with high-fidelity realism but also effectively upholds the foundational concepts, composition, stylistic elements, and visual symbolism intrinsic to genuine artworks. CreativeSynth supports a wide array of intriguing applications, including (a) image variation, (b) image editing, (c) style transfer, (d) image fusion, and (e) multimodal blending.}
}
\label{fig:teaser}
\end{figure*}

\textcolor{black}{
To address these challenges, we propose CreativeSynth, a unified framework for generating and editing artistic paintings using multimodal inputs, without requiring additional training. This framework enables users to generate and edit images that conform to personal styles and content requirements, guided by reference images and textual prompts (see Fig.~\ref{fig:teaser}).} 
\textcolor{black}{
Our work introduces a novel approach by focusing on artistic images through Cross-Art-Attention without altering the original model's parameters. This unique strategy aligns the semantic information of the natural world with the artistic paintings in a way that has not been explored in existing research. Specifically, we combine image inversion with an adaptive mechanism, which is a technical novelty that allows us to maintain the aesthetics of the original artwork at a higher level of fidelity compared to traditional methods. Through a proposed semantic fusion mechanism, we ensure that the generated images are not only simple reconstructions but also are fused and generated based on semantic and aesthetic information in multimodal inputs. In contrast, other existing methods focus on style in paintings and transfer it to the real world, lacking the comprehensive and nuanced approach we present.}

\textcolor{black}{
In this work, we begin by encoding semantic information from both images and textual prompts to establish a foundation for condition guidance.}
\textcolor{black}{
The framework then ensures aesthetic consistency through a new approach of employing shared attention to adapt the style of the semantic image to align with the target image using art-adaptive batch normalization ArtBN. This not only maintains the integrity of the artistic style but also enhances the overall visual coherence. Within the semantic fusion module, CreativeSynth leverages a decoupled cross-attention mechanism, a technical innovation that precisely coordinates interactions between visual and textual features, achieving a more cohesive and intelligent synthesis rather than a simple sum of parts. This decoupled mechanism allows for a more detailed and accurate integration of different modalities, which is a crucial contribution to the field of multimodal image generation. Furthermore, the sampling process is guided by a principle-based image inversion approach, where denoising techniques are applied to iteratively reconstruct the image from noise. This approach improves the quality and realism of the generated images, offering a more reliable solution for artistic image creation.
}
As a result, CreativeSynth generates customized artworks that faithfully reflect the provided semantic prompts and desired aesthetic style. Our method effectively handles a variety of sophisticated artistic image editing tasks, including image variation, image editing, style transfer, image fusion, and multimodal blending.
The main contributions of this paper include: 

\textcolor{black}{
\begin{itemize}
\item 
We present a novel multimodal, multitasking artistic framework that unifies diverse image editing tasks. Using a decoupled cross-attention mechanism to integrate textual and visual features, enables seamless content editing and contextual blending while preserving the original painting style. This flexibility empowers creators to modify objects, characters, or scenes freely, unlocking new possibilities for artistic expression.
\item 
By employing adaptive instance normalization to align the style of the target image with its semantic content, our approach ensures the preservation of aesthetic quality while accommodating diverse input modalities. Additionally, the framework incorporates reverse encoding with denoising before enhancing semantic control in synthesized images. These design choices facilitate the generation of high-fidelity artistic outputs, maintaining the integrity of the original artist’s style even in the presence of substantial content modifications.
\item 
Extensive experiments demonstrate that our method achieves state-of-the-art performance in visual quality while exhibiting a remarkable capability to capture style and semantic information accurately.
\end{itemize}
}

\begin{figure}
    \centering
    \includegraphics[width=\linewidth]{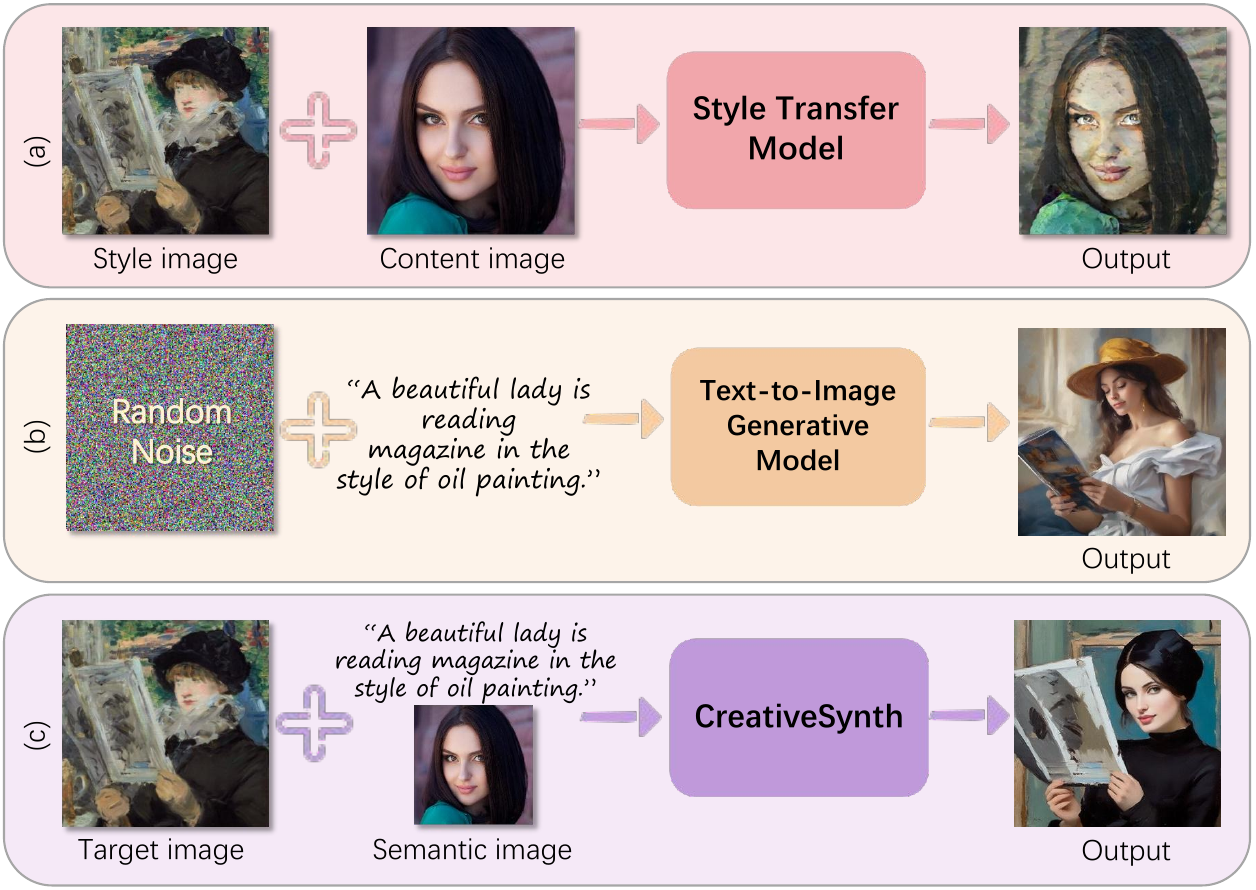}
    \caption{\textcolor{black}{The conceptual differences among the three image generation methods. (a) \textbf{Classical Style Transfer}~\cite{Huang:2017:AdaIn}, which combines a style image (providing the desired artistic style) and a content image (providing the scene or structure).
    (b) \textbf{Text-to-Image Synthesis}~\cite{sdxl}, which generates an image directly from random noise guided by textual descriptions, such as ``a beautiful lady reading a magazine in the style of oil paintings'', without requiring a reference style image.
    (c) \textbf{CreativeSynth}, which employs a ``cross-art-attention'' mechanism to seamlessly integrate semantic content with the desired style, producing outputs that are both semantically coherent and stylistically consistent.}}
    \label{fig:insight}
\end{figure}

\section{Related Work}

\subsection{\textcolor{black}{Image Style Transfer}}
Arbitrary style transfer methods~\cite{tvcg_styletransfer_review} employ unified models that handle various input styles by constructing feed-forward architectures~\cite{Liao:2017:VAT,Li:2017:UST,Park:2019:AST,deng:2021:arbitrary,lee2024audio,Deng:2024:ZST,diffstyler}. 
Huang et al.~\cite{Huang:2017:AdaIn} employ conditional instance normalization to align the overall statistics of content features with those of style features, which adjusts the statistics to achieve style transfer.
An et al.\cite{An:2021:Artflow} address content leakage through reversible neural flows and an unbiased feature transfer module called ArtFlow. 
Zhang et al.\cite{zhang2022cast} directly learn style representation from image features using contrastive learning, achieving domain-enhanced arbitrary style transfer (CAST).
In addition to CNNs, visual transformers have also been employed for style transfer tasks. 
Deng et al.\cite{deng2022stytr2} propose StyTr$^2$, a transformer-based method that considers the long-range dependencies of input images to avoid biased content representation in style transfer. 
A variety of text-based style transfer methods have emerged with the development of powerful multi-modal models.
Zhang et al.\cite{zhang2023inst} introduce InST, a diffusion stylization method based on an inversion technique that achieves more expressive style transfer. 
Sohn et al.\cite{sohn2023styledrop} propose StyleDrop, an artistic image generation method that fine-tunes a minimal number of trainable parameters within diffusion models.
While existing image style transfer methods primarily focus on learning and transferring artistic elements into a given content image (see Fig.~\ref{fig:insight}(a)), our approach aims to create the appearance of specific content within a target painting.

\begin{figure*}
    \centering
    \includegraphics[width=0.9\linewidth]{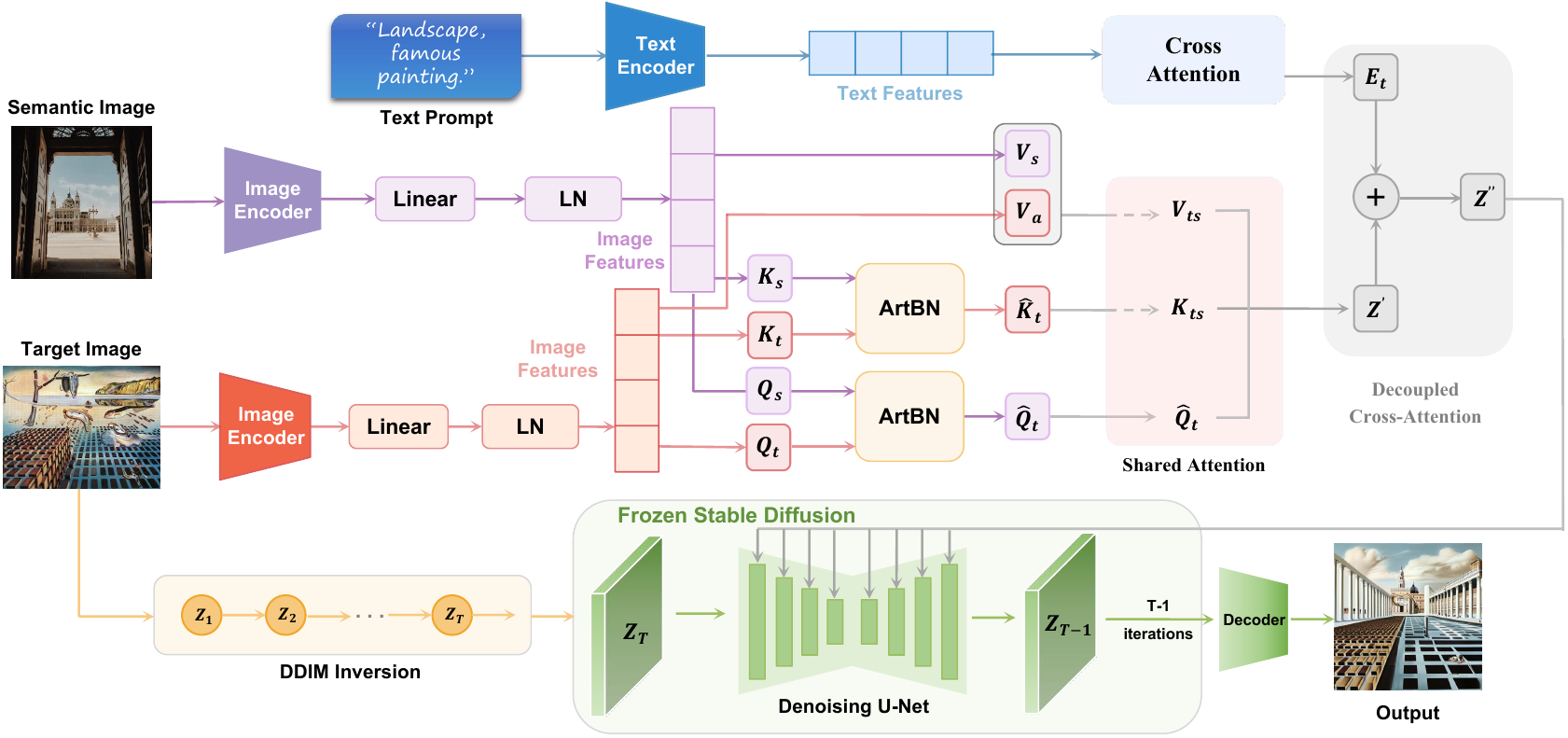}
    \caption{\textcolor{black}{The overall structure of CreativeSynth. Text features and image features are first acquired from separate text and image encoders, respectively. Then, target and semantic images are interacted by applying AdaIN to focus on image art features. An innovative decoupled cross-attention mechanism is employed to fuse the attention between the multimodal inputs, which is subsequently integrated into a U-Net architecture. The target image is transformed into a latent variable $z_T$ via DDIM Inversion, and the final output is refined through a denoising network.}}
    \label{fig:pipeline}
\end{figure*}

\subsection{\textcolor{black}{Text-to-image Generation}}
With the ability of neural networks to understand intricate natural language and visual representations, the field of image synthesis has made significant progress from textual descriptions~\cite{devlin2018bert}. 
Transformer-based architectures such as DALL-E~\cite{dalle} and its follow-up studies~\cite{dalle2,glide} incorporate powerful attentional mechanisms to efficiently transform textual prompts into high-fidelity images. Similarly, VQ-VAE-2~\cite{VQ-VAE-2} and its autoregressive model demonstrate the strong potential of combining textual and visual patterns through discrete latent spaces. These methods have achieved remarkable results, but they often do not allow for fine control of structural details~\cite{stylealigned,masactrl}. Diffusion models similar to Stable Diffusion~\cite{latentdiffusion} also exemplify the ability to generate high-quality images based on descriptions. Nonetheless, as shown in Fig.~\ref{fig:insight}(b), these methods still face the challenge of generating images with styles that are inconsistent with textual prompts. Our research closely follows the previous work~\cite{mgad,dreambooth,TI,stylealigned}, focusing on converting multimodal prompts into realistic artistic images and achieving innovations in reconstructing and editing existing images.

\subsection{\textcolor{black}{Personalized Image Generation}}
In order to incorporate specific styles or personalities into image generation, personalization, and style alignment has become an important area of research. For example, StyleGAN~\cite{stylegan} has made impressive progress in personalized face generation.
ControlNet~\cite{ControlNet} leverages ``zero-convolution'' fine-tuning on pre-trained diffusion models to enable diverse, prompt-driven image generation with spatial conditioning.
In terms of image restoration with constraints, ProSpect~\cite{prospect} attempts to preserve the style features of the semantic image while adapting its content to fit the new context. In terms of achieving multi-image style consistency, Style Aligned~\cite{stylealigned} shows how multiple images can be stylistically consistent through a shared attention layer. Textual Inversion~\cite{TI} introduces a method for embedding new "words" into the model space using as few as 3-5 images, allowing for nuanced linguistic guidance customization and demonstrating superior concept depiction capabilities across a range of tasks. 
\cite{ddimInversion} enables intuitive text-guided image edits by inverting images into the domain of a pre-trained model using meaningful text prompts.
As demonstrated in Fig.~\ref{fig:insight}(c), our work extends the above idea by enhancing the interaction between textual and artistic visual features made achievable through multimodal fusion.

\section{Method}

\subsection{Overview}
\textcolor{black}{
CreativeSynth synthesizes text and image data to produce artwork guided by specific conditions, leveraging our core innovation, the cross-art-attention mechanism (see Sections~\ref{3.3}-\ref{3.5}). This approach establishes novel connections between diverse art-related information, aligning textual art descriptions with visual art elements at both semantic and stylistic levels, thereby enhancing feature integration.
Cross-art-attention is pivotal, intertwining artistic concepts, emotions, visual components, and compositional styles of the target images with the semantic content requirements from text prompts and source images. This integration enriches the semantic depth of the input and yields artistic outputs that are detailed, accurate, and consistent in style and meaning.
As depicted in Fig.~\ref{fig:pipeline}, this approach initially conducts semantic encoding of the semantic image and text prompts, providing content information that lays the groundwork for condition guidance. Subsequently, our framework focuses on aesthetic preservation, where a dedicated processor adjusts the style of the semantic image to align with that of the target image through adaptive instance normalization, ensuring that the generated output maintains stylistic consistency with the desired aesthetic.
During the semantic fusion phase, a decoupled cross-attention mechanism harmonizes visual and textual features into a unified representation.
The sampling process is guided by the image inversion principle, which leverages denoising to reconstruct the image from noise. As a result, CreativeSynth generates bespoke artworks that faithfully reflect the provided semantic conditions and aesthetic preferences.
}

\subsection{Prelimiaries}
The diffusion process is simulated through a gradual noise addition process, where noise is progressively introduced to the clear original image $x_0$, generating a series of transitional latent variables $(x_1, ..., x_T)$. In the denoising diffusion model, this process is defined as:
\begin{equation}
x_{t} = \sqrt{\bar{\alpha}_{t}} x_{0}+ \sqrt{1 - \bar{\alpha}_{t}} \epsilon_{\theta}(x_{t-1}, t),  
\end{equation}
where $\bar{\alpha}_{t} = \prod_{i=1}^{t} \alpha_{i}$ is the cumulative product factor for time step t, and $\epsilon$ is a neural network model learned from random noise. In this process, we gradually apply forward noising to the original image $x_0$, creating a series of increasingly noised images $x_1, x_2, \ldots, x_T$ until an image that is nearly pure noise $x_T$ is generated. Subsequently, we reconstruct the image using a reverse process, that is, by denoising steps learned step by step, we trace back from the noisy image $x_T$ to the original image $x_0$. The key to our approach is the Denoising Diffusion Implicit Models (DDIMs)~\cite{ddim}, which enables precise control over synthesis, serving as the backbone of the algorithm. DDIM employs a non-Markovian diffusion process, characterized by a sequence of forward noising steps followed by a reverse denoising procedure.

In order to reconstruct a real image under a given conditional text, we need to perform a reverse process to recover the image from random noise. We employ the deterministic DDIMs as our core denoising technique. Specifically, we use the following reverse formula of DDIM to restore the original image:
\begin{equation}
x_{t-1} = \frac{1}{\sqrt{\alpha_{t}}} \left(x_{t} - \frac{1 - \alpha_{t}}{\sqrt{1 - \bar{\alpha}_{t}}} \epsilon_{\theta}(x_{t}, t) \right) + \sigma_{t} \beta,  
\end{equation}
where $\alpha_{t}$ is the step size factor in the denoising process, $\epsilon_{\theta}$ is the predicted noise, $\beta$ is an optional noise vector used to increase randomness, and $\sigma_{t}$ is a factor that regulates the noise intensity.

\subsection{Condition Guidance}
\label{3.3}
The encoding process integrates text and image features using a decoupled cross-attention mechanism within the framework of a pre-trained Stable Diffusion model~\cite{latentdiffusion}.
For a given text prompt $P$, the tokenizer and the text encoder from the pre-trained diffusion model are used to generate the text embeddings $\mathbf{E}_{\text{text}} \in \mathbb{R}^{n \times d_{\text{text}}}$:
\begin{equation}
\mathbf{E}_{\text{text}} = \mathcal{E}(\text{Tokenizer}(P)),
\end{equation}
where $n$ is the sequence length and $d_{\text{text}}$ is the text embedding dimension.

\textcolor{black}{The image encoder $\mathcal{E}_{\text{I}}$ is trained using the IP-Adapter framework~\cite{ip-adapter} built upon the Stable Diffusion XL-1.0 model~\cite{sdxl}. This encoder transforms input images into latent representations optimized for processing by the generative model.
Specifically, for an input image $\mathbf{I}$, the latent encoding is obtained through a forward pass within the image encoder network:
\begin{equation}
\mathbf{z}_{I} = \mathcal{E}_{\text{I}}(\mathbf{I}). 
\end{equation}
}
Unlike the existing U-Net cross-attention architecture, which uses two paths to process text and image features separately, each path consists of specialized cross-attention layers that are dedicated to either text features or image features without interfering with each other.
Instead, we use a decoupled cross-attention mechanism, where shared attention results for images and cross-attention results for text are combined for final image generation.

\begin{figure*}
    \centering
    \includegraphics[width=\linewidth]{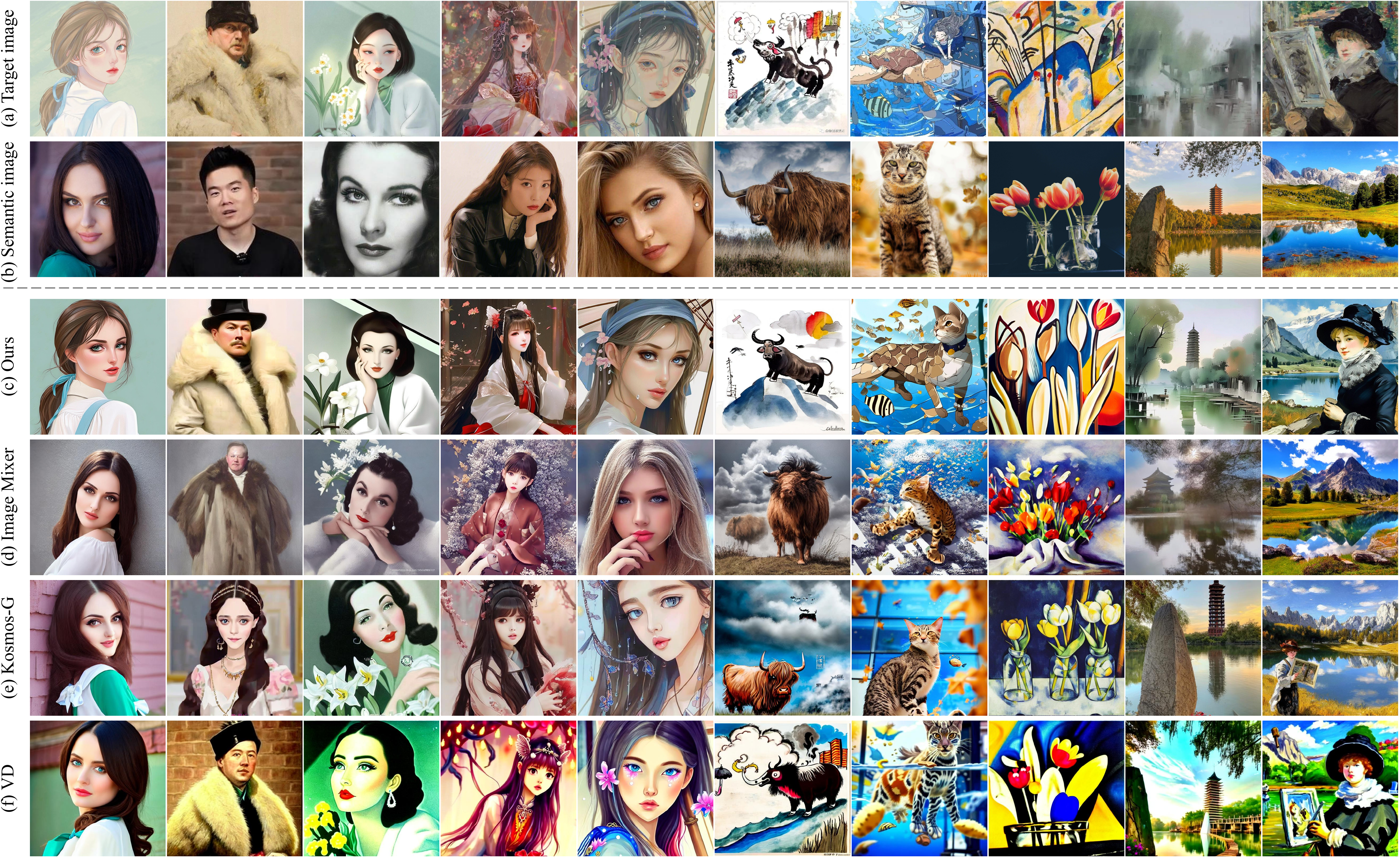}
    \caption{Qualitative comparisons of our proposed CreativeSynth with other extant methods. The results offer a visualization of image fusion between artistic and real images. }
    \label{fig:comparison_i+i}
\end{figure*}

\subsection{Aesthetic Maintenance}
\label{3.4}
\textcolor{black}{The Aesthetic Maintenance module is designed to bridge the gap between the generated images and the art world's aesthetic standards, which are characterized by finer-grained dimensions such as color, technique, and composition. The module enables the model to prioritize key visual features and ensure overall harmony, by leveraging style alignment and shared attention mechanisms. Through these mechanisms, the model inherits accurate painting techniques and constructs visually balanced compositions.}

\subsubsection{\textcolor{black}{Self-attention} }
Leading text-to-image (T2I) diffusion models~\cite{sdxl,latentdiffusion,ControlNet} integrate a U-Net structure~\cite{unet} composed of convolutional layers and transformer-based attention blocks~\cite{Attentionisall}. Within these attention frameworks, deep image features $\phi \in \mathbb{R}^{m \times d_h}$ undergo self-attention within layers and interact with contextual text embeddings through cross-attention layers.

In our study, we focus on the self-attention layers, where the deep features are updated by mutually attending to one another. Initially, the features are transformed into query matrices $Q \in {m \times d_k}$, key matrices $K \in {m \times d_k}$, and value matrices $V \in {m \times d_h}$ via learned linear transformations. Following this, the attention mechanism is calculated using scaled dot-product attention:
\begin{align}
\textrm{Attention}(Q, K, V) = \textrm{softmax}\left(\frac{QK^T}{\sqrt{d_k}}\right) V,
\end{align}
where $d_k$ denotes the dimensionality of $Q$ and $K$. Conceptually, the update for each image feature is a weighted sum of $V$, with weights derived from the correlation between the projected queries $q$ and the keys $K$.

\subsubsection{Style alignment}
\textcolor{black}{
To achieve stylistic harmony between images, it is essential to facilitate the flow of attention from the semantic image to the target image. The style alignment module employs designed art batch normalization (ArtBN) to align the queries and keys of both the semantic and target images to a specified style. This approach ensures that the generated images not only maintain stylistic coherence with the desired aesthetic but also enhance visual appeal and thematic consistency within the synthesized content.
By receiving target and semantic inputs, the channel mean and variance of the semantic input are adjusted to align with those of the target input. The ArtBN definition is as follows:}
\textcolor{black}{
\begin{equation}
\hat{Q}_t=\operatorname{ArtBN}\left(Q_t, Q_s\right),\quad \hat{K}_t=\operatorname{ArtBN}\left(K_t, K_s\right),
\end{equation}
}
where ${Q}_s$ and ${K}_s$ are the query and key of the semantic image, and ${Q}_t$ and ${K}_t$ are the query and key of the target image, respectively. 
\textcolor{black}{
When only receiving target input, for image variation and editing applications, the ArtBN definition is as follows:
\begin{equation}
\operatorname{ArtBN}(x)=\frac{x-\mu(x)}{\sigma(x)}.
\end{equation}
}
For an input batch $x \in \mathbb{R}^{N \times C \times H \times W}$, $\mu(x)$ and $\sigma(x)$ are computed across spatial dimensions independently for each channel and each sample:
\begin{equation}
\mu_{n c}(x)=\frac{1}{H W} \sum_{h=1}^H \sum_{w=1}^W x_{n c h w},
\end{equation}
\begin{equation}
\sigma_{n c}(x)=\sqrt{\frac{1}{H W} \sum_{h=1}^H \sum_{w=1}^W\left(x_{n c h w}-\mu_{n c}(x)\right)^2+\epsilon}.
\end{equation}

\subsubsection{Shared attention.} 
\textcolor{black}{
Shared attention integrates the characteristics of both the target and semantic images, updating the information in the semantic images based on the style of the target image. ${K}_{ts}$ represents the shared key, respectively, while ${V}_{ts}$ denotes the value:
\begin{equation}
{K}_{ts} = \begin{bmatrix} {K}_s \\ {\hat{K}}_t \end{bmatrix},\quad
{V}_{ts} = \begin{bmatrix} {V}_s \\ {V}_t \end{bmatrix}.
\end{equation}
When only the target input is received,
\begin{equation}
{K}_{ts} = {\hat{K}}_t ,\quad
{V}_{ts} = {V}_t.
\end{equation}
The keys and values are jointly aggregated from both the target image and the semantic image, whereas the query exclusively represents the attributes of the target image. The scaled dot-product attention mechanism is then applied as follows:
\begin{equation}
\mathbf{Z}' = \text{Attention}(\hat{{Q}}_t, {K}_{ts}^T, {V}_{ts}) = \text{Softmax}\left(\frac{\hat{{Q}}_t {K}_{ts}^{\top}}{\sqrt{d}}\right){V}_{ts},
\end{equation}
where ${\hat{Q}}_t$ represent the query normalized by ArtBN and $d$ is the dimensionality of the keys and queries.
}

\subsection{\textcolor{black}{Semantic Fusion}}
\label{3.5}
\textcolor{black}{The semantic fusion module facilitates multi-level interactions between semantic images, target images, and text by employing a decoupled cross-attention mechanism. This mechanism processes each information stream (image and text features) through separate cross-attention layers, enabling iterative co-optimization of cross-modal information. Thus the module significantly enhances the model's capacity to understand and generate outputs for complex art creation and analysis tasks. Ultimately, the information streams are fused to generate the final modified image features $\mathbf{Z}''$:}
\begin{equation}
\mathbf{Z}''=\mathbf{Z}'+ \text{Softmax}\left(\frac{{Q}{K}^{\top}}{\sqrt{d}}\right){V},
\end{equation}
where ${Q}, {K}, {V}$ are the transformed query, key, and value matrices of the text features. The contributions of individual decoupled attention operations are aggregated to shape the final feature representation.

\subsection{Sample Process}
In our method, the reverse diffusion process adheres to the formal representation below to progressively restore the latent clean image:
\begin{align}
\mathbf{z}_{t-1} &= \sqrt{\alpha_{t-1}} \left(\frac{\mathbf{z}_t - \sqrt{1-\alpha_t} \epsilon_{\theta}(\mathbf{z}_t,t)}{\sqrt{\alpha_t}}\right) \notag \\
&+ \sqrt{1 - \alpha_{t-1}} \epsilon_{\theta}(\mathbf{z}_t,t),
\end{align}
where $\alpha_t$ represents the steps of the predetermined variance schedule, and $\epsilon{\theta}$ is a parameterized neural network responsible for predicting the noise component in the image at time $t$. The process begins with the initial latent representation, $\mathbf{z}_T$, sampled from a prior noise distribution, $\mathcal{N}(0, \mathbf{I})$, and progressively reduces the noise to recover the original image representation in the latent space, $\mathbf{z}_0$. This gradual denoising enables our model to achieve precise image synthesis while ensuring robust control and predictability throughout the synthesis process.

In the sampling process, we design an inversion callback function to adjust the latent space vectors at the end of each inversion step, ensuring the text alignment of the image. This function is defined as:
\begin{equation}
\mathrm{Callback}(z_t, t) = 
\begin{cases}
z_T, & \text{if } t = T, \\
z_t, & \text{otherwise},
\end{cases} 
\end{equation}
where $z_t$ denotes the latent variable corresponding to the temporal index $t$, which is replaced with a pre-computed vector derived via the DDIM inversion~\cite{dhariwal2021diffusion}. This ensures that throughout the diffusion process, our optimized latent space vector remains highly consistent with the inherent attributes of the target image.

\begin{figure*}
    \centering
    \includegraphics[width=\linewidth]{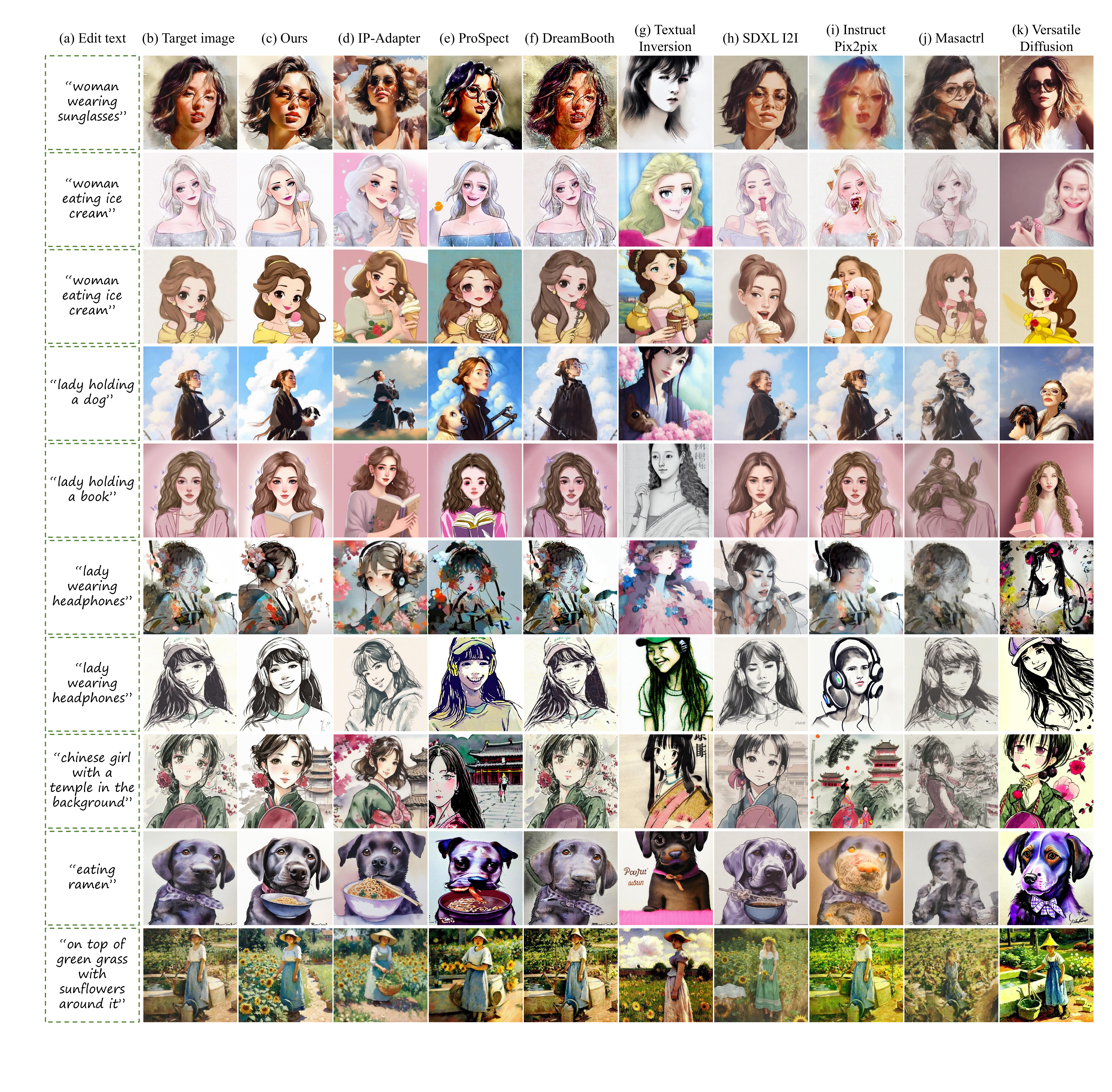}
    \caption{\textcolor{black}{Visual comparison of our proposed CreativeSynth with state-of-the-art methods for text-guided editing of diverse types of art images. }}
    \label{fig:comparison_i+t}
\end{figure*}

\section{Experiments}
\subsection{Implementation Details}
The Stable Diffusion XL (SDXL)~\cite{sdxl} model utilized in our experiments has been pre-trained on a large-scale corpus of text-image pairs, fully leveraging its representational capacity. To maintain experimental consistency, we standardized the number of generation steps to 30 and the guidance scale to 5.0 throughout our evaluations. Furthermore, the input images utilized in the experiments are uniformly scaled to a resolution of \ensuremath{1024 \times 1024} pixels for both the image reversal process and subsequent image synthesis tasks. \textcolor{black}{The generation of each image requires approximately five seconds on a single NVIDIA L40 GPU and around two seconds on a single NVIDIA RTX 4090 GPU.}

\begin{table*}[!h]
  \centering
  \fontsize{7pt}{10pt}\selectfont
  \caption{\textcolor{black}{Statistics of quantitative comparison with state-of-the-art methods for text-guided image editing. Specific metrics include Aesthetic Score, NIMA, CLIP-T, and CLIP-I. The best results are in \textbf{bold} while the second-best results are marked with \underline{underline}. }}
    \resizebox{\textwidth}{!}{\begin{tabular}{l|ccccccccccc}
\toprule
    & Ours   & IP-Adapter     & ProSpect      & DreamBooth       & TI    & SDXL I2I    & Instruct P2P     & Masactrl      & VD        \\
\midrule
    Aesthetic Score $\uparrow$ &\textbf{7.563}   &  \underline{7.249}  & 6.297       & 6.216      & 6.441
       & 6.636    & 5.344   & 5.707      & 6.818   
 \\
NIMA $\uparrow$ &\textbf{6.4382}   &  \underline{5.9713}  & 4.6754       & 5.3396      & 4.5942
   & 5.3647    & 3.7937   & 3.9487      & 4.3901  
\\
    CLIP-T $\uparrow$&\textbf{59.123} &  57.956 &\underline{58.004} &46.792 &48.576 &57.981 &55.203 &45.147 &53.516 \\
    CLIP-I $\uparrow$&\underline{69.84}  &54.62   &58.79 &\textbf{84.36} &39.31 &63.14 &48.41 &51.95 &45.46 \\
\bottomrule
    \end{tabular}}
  \label{tab:tab_quantity1}%
\end{table*}%

\begin{table*}[htbp]
  \centering
  \caption{\textcolor{black}{Statistics of quantitative comparison with state-of-the-art methods for Image fusion. Specific metrics include Aesthetic Score, NIMA, and CLIP-I. The best results are in \textbf{bold} while the second-best results are marked with \underline{underline}.}}
  \setlength{\extrarowheight}{2.2pt}
  \resizebox{0.5\textwidth}{!}{ \begin{tabular}{l|ccccccccccc}
\toprule
    & Ours    & VD       & Image Mixer       & Kosmos-G \\
\midrule
    Aesthetic Score $\uparrow$ &\textbf{7.563}     & 6.818   & \underline{7.151}       & 6.125 \\
    NIMA $\uparrow$ &\textbf{6.3648}      & 4.4269  & \underline{5.3955}      & 5.3071 \\
    CLIP-I $\uparrow$&\textbf{52.067}  &44.973 &48.349 &\underline{50.564}\\
\bottomrule
    \end{tabular}}
  \label{tab:tab_quantity2}%
\end{table*}%

\subsection{Qualitative Evaluation}
\subsubsection{Image fusion}

We conduct comparisons for image fusion tasks with state-of-the-arts methods including Image Mixer~\cite{imagemixer}, Kosmos-G~\cite{Kosmos-g}, and Versatile Diffusion~\cite{vd}. Qualitative results are shown in Fig.~\ref{fig:comparison_i+i}.
Image Mixer and Kosmos-G tend to generate results with subdued stylistic expression, often producing images that are more realistic than artistic.
Versatile Diffusion, by contrast, consistently demonstrates strong creative expressiveness across a variety of outputs but struggles to capture the nuanced subtleties of distinct styles. In comparison, our approach consistently produces harmoniously fused results that balance creative excellence and aesthetic appeal, while excelling in style representation and effectively integrating semantic information.

\subsubsection{Text guided image editing}
To accurately assess model performance, we conduct baseline comparisons for the task of single-image text editing. As shown in Fig.~\ref{fig:comparison_i+t}, our model takes succinct personalized text descriptions as input and successfully performs operations such as semantic introduction, facial attribute modification, and complex scene recreation across a range of different scenes and objects. For a comprehensive evaluation of our method, we select several advanced baseline models for comparison, including IP-Adapter~\cite{ip-adapter}, ProSpect~\cite{prospect}, DreamBooth~\cite{dreambooth}, Textual Inversion~\cite{TI}, SDXL I2I~\cite{sdxl}, Instruct Pix2Pix~\cite{instructp2p}, Masactrl~\cite{masactrl}, and Versatile Diffusion (VD)~\cite{vd}. 

Based on the results, although the IP-Adapter generates results of superior quality, it fails to preserve the non-editing information of the target image.
In terms of style consistency, some models like ProSpect, DreamBooth, and SDXL I2I exhibit high congruence with the target image. However, Instruct Pix2Pix and Masactrl often damage the composition and content of the target image during editing, introducing distortions and unnatural artifacts. For instance, images processed by Instruct Pix2Pix show obvious ghosting effects on headphones and ice cream, while Masactrl faces challenges in generating specific and realistic human faces. ProSpect and SDXL I2I perform admirably in preserving the semantic content of the original image, but often experience significant alterations in key features such as facial details of people, impacting the image's authenticity and credibility. In contrast, DreamBooth's results display very limited input image information changes, leading to the production of images that nearly do not align with the text editing requirements, thus limiting their potential for practical application. Lastly, Textual Inversion and Versatile Diffusion can generate quite distinctive artworks, which are creative but still deviate significantly from the target image in terms of style and semantic preservation.

Compared to baselines, our results guarantee a high level of content fidelity and stylistic coherence during image modification. The altered images retain the principal structure of the original while integrating new attributes or alterations in accordance with text-based directives. In the domain of facial attribute editing, our method yields facial features that are both more natural and realistic, minimizing visual anomalies and undue artistic alterations. Furthermore, our approach facilitates the effective reconstruction and editing of intricate scenes without disrupting the global composition of the image.

\subsection{Quantitative Evaluation}
\textcolor{black}{To comprehensively assess the performance of our proposed method, this paper employs four key metrics—Aesthetic Score~\cite{aestheticscore}, NIMA~\cite{nima}, CLIP-T~\cite{clip}, and CLIP-I~\cite{clip}—for quantitative comparison with the state-of-the-art methods.
The Aesthetic Score reflects the visual appeal and artistic quality of the generated images, NIMA evaluates the realism and naturalness of the images, CLIP-T describes the semantic consistency between the generated images and the edited text, and CLIP-I indicates the visual and content coherence between the generated images and the target images.}

\textcolor{black}{
\textbf{Text guided image editing.}
The comparison results are shown in Table~\ref{tab:tab_quantity1}. In terms of Aesthetic Score, our method significantly outperforms other methods, achieving the highest average score of $7.563$, demonstrating its excellent performance in overall aesthetic appeal. In the NIMA score, our method also achieves a high score of $6.4382$, indicating that our generated images excel in realism and naturalness. On the CLIP-T metric, our method leads with a score of $59.123$, which shows the effectiveness of our method in ensuring semantic consistency between the generated images and text descriptions. Furthermore, our method also achieves a high score of $69.84$ on the CLIP-I metric, highlighting our method's ability to maintain visual semantics and detail fidelity. While Dreambooth achieved the highest score in the benchmarks, it struggled when executing text commands.}

\textcolor{black}{
\textbf{Image fusion.}
The comparison results are shown in Table~\ref{tab:tab_quantity2}. In terms of Aesthetic Score, our method also performs exceptionally well, scoring $7.563$, which is a high score in the task of image fusion. In the NIMA score, our method scores $6.3648$, which shows the advantage of our image fusion results in terms of realism and naturalness. On the CLIP-I metric, our method scores $52.067$, indicating that our method can effectively maintain visual quality and content consistency when it comes to image fusion.}

\begin{figure*}[!h]
    \centering
    \includegraphics[width=\textwidth]{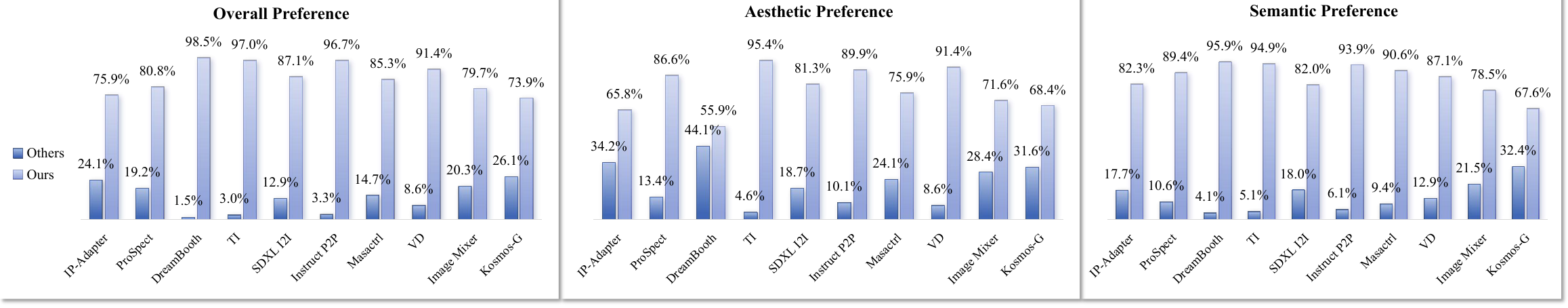}
    \caption{\textcolor{black}{The user study with the comparative method consisted of three aspects: overall preference, aesthetic preference, and semantic preference. Users chose between the comparative methods and CreativeSynth. The table shows the percentage of people who chose the comparison method.}
     }
    
    \label{fig:preference}
\end{figure*}
\begin{table*}[htbp]
  \centering
  \caption{\textcolor{black}{The user study with the comparative method consisted of three aspects: overall preference, aesthetic preference, and semantic preference. Users chose between the comparative methods and CreativeSynth. The table shows the percentage of people who chose the comparison method.}}
  \setlength{\extrarowheight}{2.3pt}
    \resizebox{\textwidth}{!}{\begin{tabular}{l|cccccccccc}
\toprule
      & IP-Adapter     & ProSpect      & DreamBooth       & TI    & SDXL I2I    & Instruct P2P     & Masactrl      & VD       & Image Mixer       & Kosmos-G \\
\midrule
    Overall Preference   & 24.1\%  &19.2\% &1.5\% &3.0\% &12.9\% &3.3\% &14.7\% &8.6\% &20.3\% &26.1\% \\
    Aesthetic Preference  & 34.2\%  &13.4\% &44.1\% &4.6\% &18.7\% &10.1\% &24.1\% &8.6\% &28.4\% &31.6\% \\
    Semantic Preference   & 17.7\%  &10.6\% &4.1\% &5.1\% &18.0\% &6.1\% &9.4\% &12.9\% &21.5\% &32.4\%  
    \\
\bottomrule
    \end{tabular}}
  \label{tab:userstudy}%
\end{table*}%
\subsection{User study}
\subsubsection{Preference}
We benchmark CreativeSynth with ten other leading-edge image-generation techniques to determine which generates the most favored artistic outcomes. We presented each participant with $50$ randomly selected sets of results, displaying the images produced by CreativeSynth and an alternative method in no particular order.
We asked participants to identify the results that (1) were the most visually pleasing overall, (2) most closely matched the artistic expression of the target image, and (3) most closely related to the editorial semantics of the text or image.
In the end, we obtained $11,850$ votes from $79$ participants, and the percentage of votes for each method is detailed in Fig.~\ref{fig:preference}. It is worth noting that CreativeSynth is particularly popular in the categories of ink drawing, oil painting, and digital art.

\subsubsection{Necessity study}
We also included the following inquiries in the user research questionnaire to help determine the requirement of our suggested idea: (1) does it generate realistic artwork?; (2) is the outcome innovative and interesting?; and (3) is it necessary?
Finally, we collected $237$ votes from $79$ participants. The voting results showed that a percentage of $94.9\%$ supported the necessity of our idea, the potential of the idea to be applied in practice, and the possibility of fulfilling a specific need. In addition, $91.1\%$ of the participants believed that it is indeed possible to create highly realistic works of art based on our idea, which shows the credibility of our technical realization and the recognition of the expected results. Most strikingly, a whopping $96.2\%$ found the artworks generated by our idea to be innovative and appealing. In view of this, our idea is not only widely recognized by the potential user community, but also has some prospects for practical application.

\begin{figure}
    \centering
    \includegraphics[width=\linewidth]{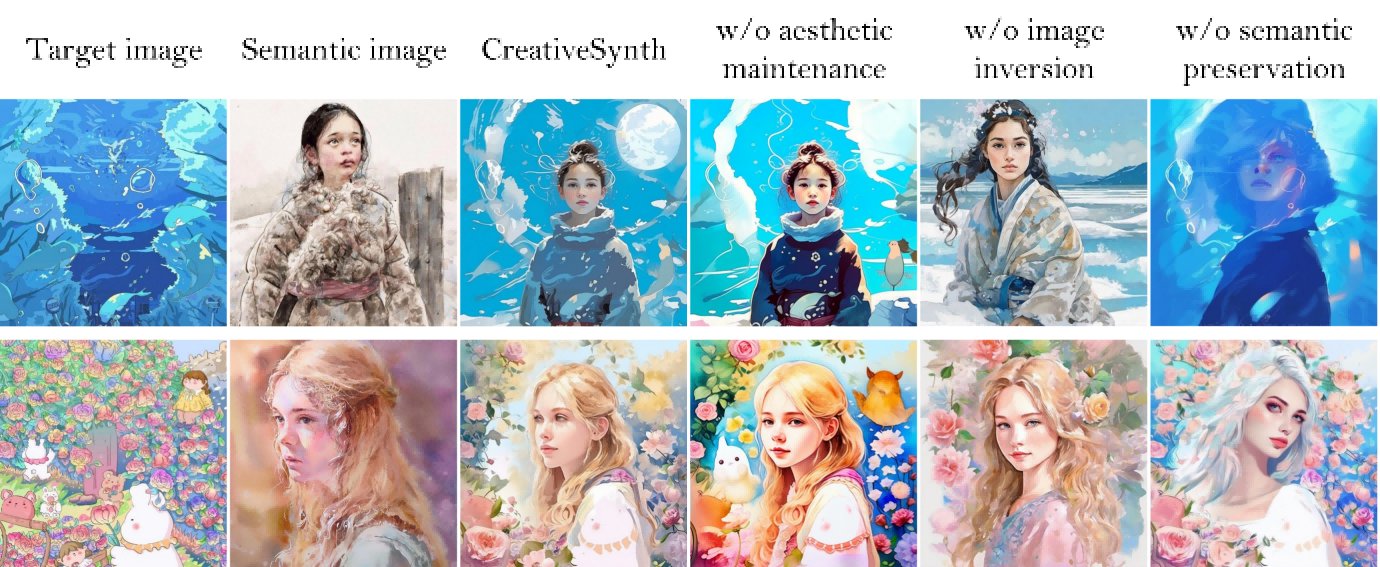}
    \caption{Results of ablation study on mechanism dissection. The \textcolor{black}{aesthetic maintenance}, image inversion, and semantic preservation components play key roles in style maintenance, detail consistency and semantic preservation respectively.}
    \label{fig:ablation_Mechanism_dissection}
\end{figure}

\begin{figure}
    \centering
    \includegraphics[width=\linewidth]{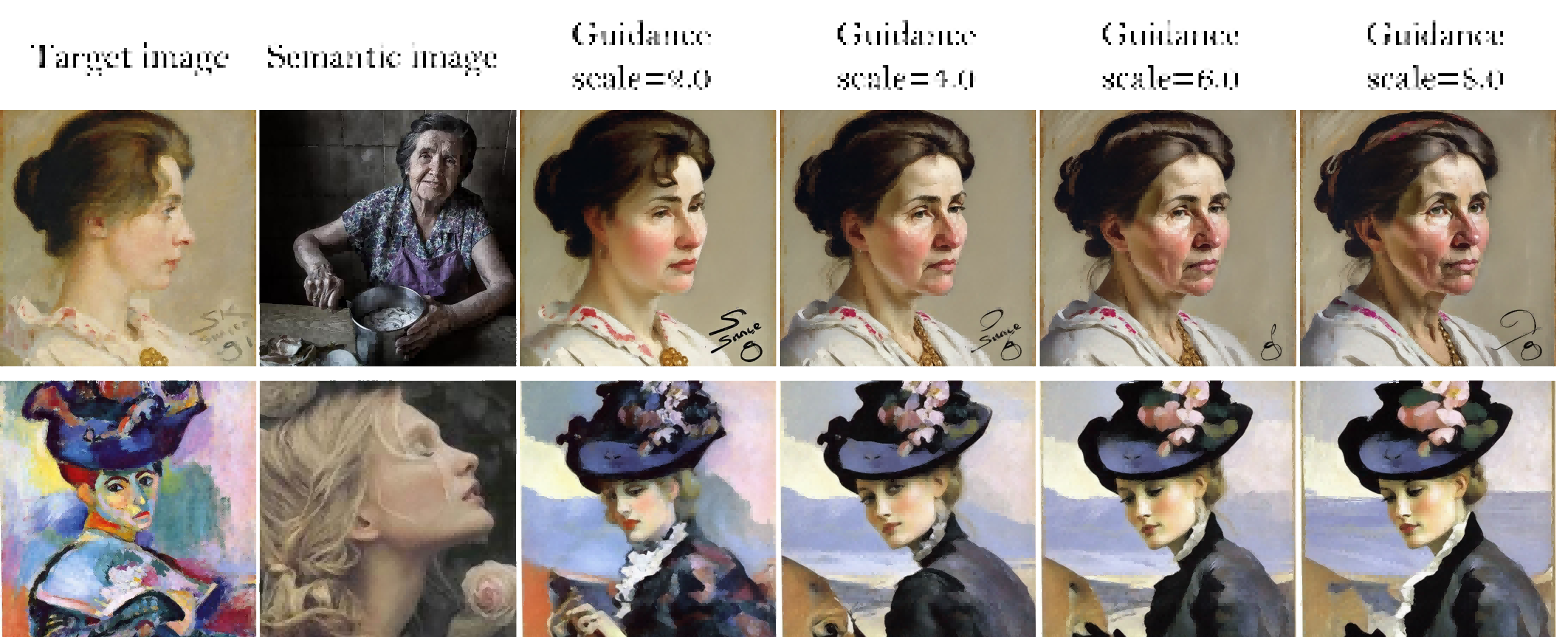}
    \caption{Results of ablation study on semantic preservation guidance scale. As the scale increases, the degree of semantic image guidance increases and the expression of semantic information in the results is enhanced.}
    \label{fig:ablation_Semantic_preservation}
\end{figure}

\begin{figure}
    \centering
    \includegraphics[width=\linewidth]{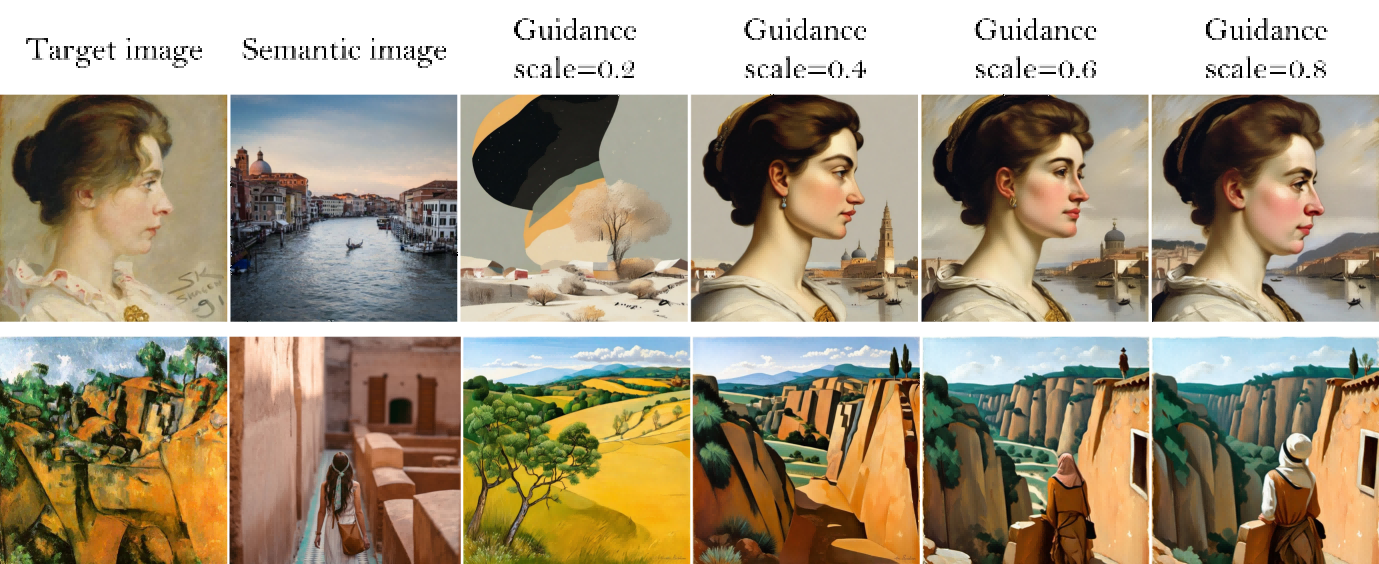}
    \caption{Results of ablation study on condition guidance scale. As the condition guidance scale increases, the agreement between the results and the target image increases.}
    \label{fig:ablation_Condition_guidance}
\end{figure}

\subsection{Ablation Study}
To evaluate the effectiveness of the various components in CreativeSynth, we conducted an ablation study incorporating the concepts of \textcolor{black}{aesthetic maintenance}, image inversion, semantic preservation, and condition guidance.

\subsubsection{Mechanism dissection} 
We show the ablation results for the main mechanisms in Fig.~\ref{fig:ablation_Mechanism_dissection}. The third column of the figure shows the results of the full model output. The fourth to sixth columns, on the other hand, show the images generated without style understanding, image inversion, and semantic preservation, respectively. 
The results show that when the model includes the style understanding component, it is able to effectively reproduce the stylistic features of the input images and generate stylistically consistent and aesthetically pleasing output images. In contrast, models lacking style understanding produce images with oversaturated colors.
The image inversion component enhances the model's understanding by parsing the original image details. This mechanism allows the generated image to better retain the detailed information of the original image, improving visual coherence and realism.
The inclusion of semantic preservation helps to generate a more rational structure and more complete semantic information, highlighting the effectiveness of semantic preservation in improving overall quality.

\subsubsection{Semantic preservation}
As shown in Fig.~\ref{fig:ablation_Semantic_preservation}, we adjusted the guidance scale of the semantic images and recorded the changes in the generated images at different scales. As the guidance scale increases, the generated images reproduce the input semantic information more accurately under the given conditions, producing well-structured and semantically rich images. 
On the other hand, the generated outputs fail to capture the essential information of the input and instead create confused semantic information at lower guiding scales.
In addition, our method generates images with more harmonious tones, textures, and lighting between the foreground and background, resulting in more realistic artistic results.

\subsubsection{Condition guidance}
By analyzing the visual results provided in Fig.~\ref{fig:ablation_Condition_guidance}, we discover that as the guidance scale increases, the details of the results become richer and more precise, aligning closer to the target style. This set of experiments, supported by both the visual demonstration in Fig.~\ref{fig:ablation_Condition_guidance}, confirms that increasing the guidance scale significantly improves the clarity and detail representation of the images, as well as the controllability over the generated images, thereby enhancing their accuracy and editability. Consequently, adjusting the guidance scale parameter effectively optimizes the performance of our image generation algorithm.

\begin{figure}
    \centering
    \includegraphics[width=\linewidth]{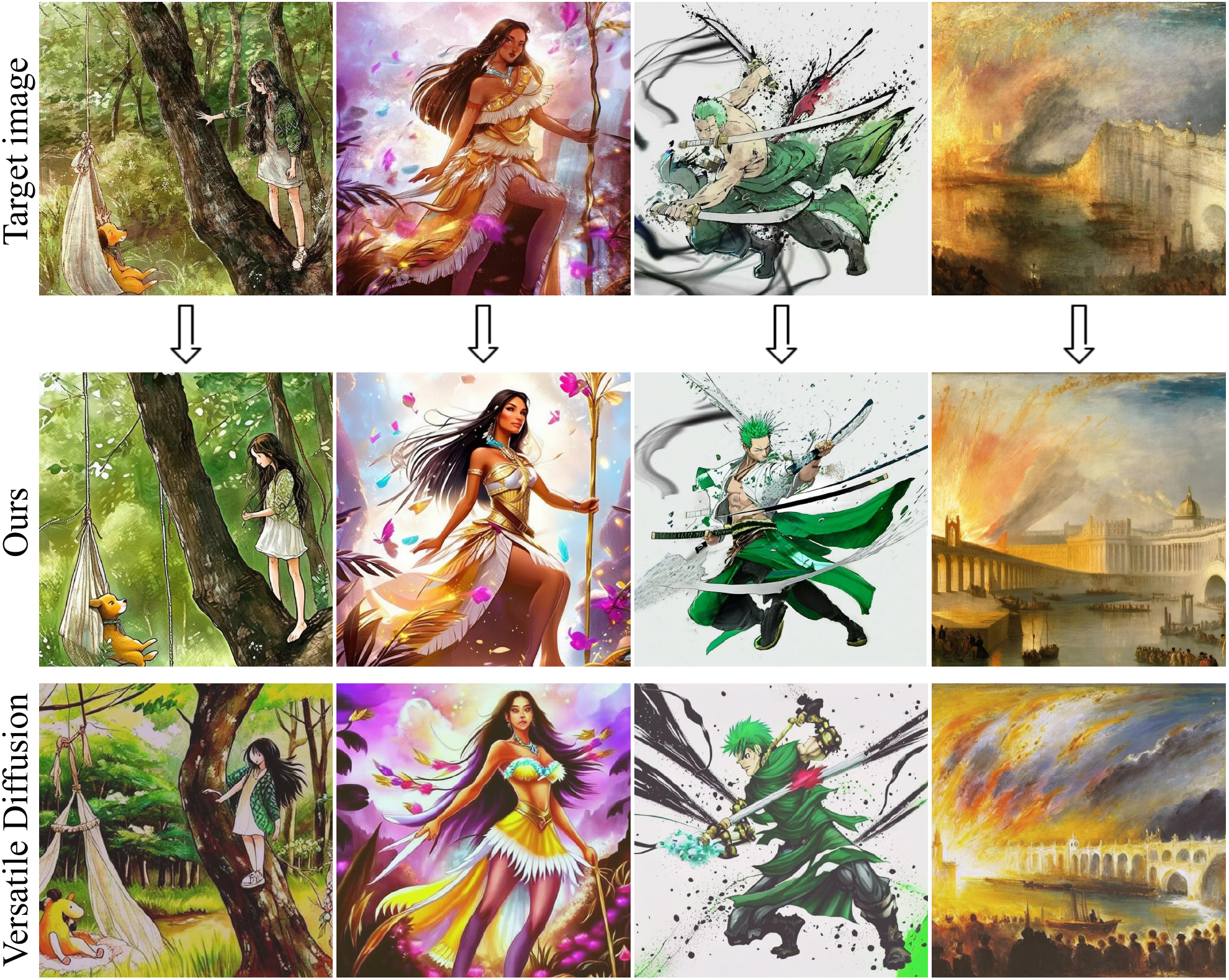}
    \caption{\textcolor{black}{Implement the display of the results of Image variation by using CreativeSynth.}}
    \label{fig:variation}
\end{figure}

\begin{figure}
    \centering
    \includegraphics[width=\linewidth]{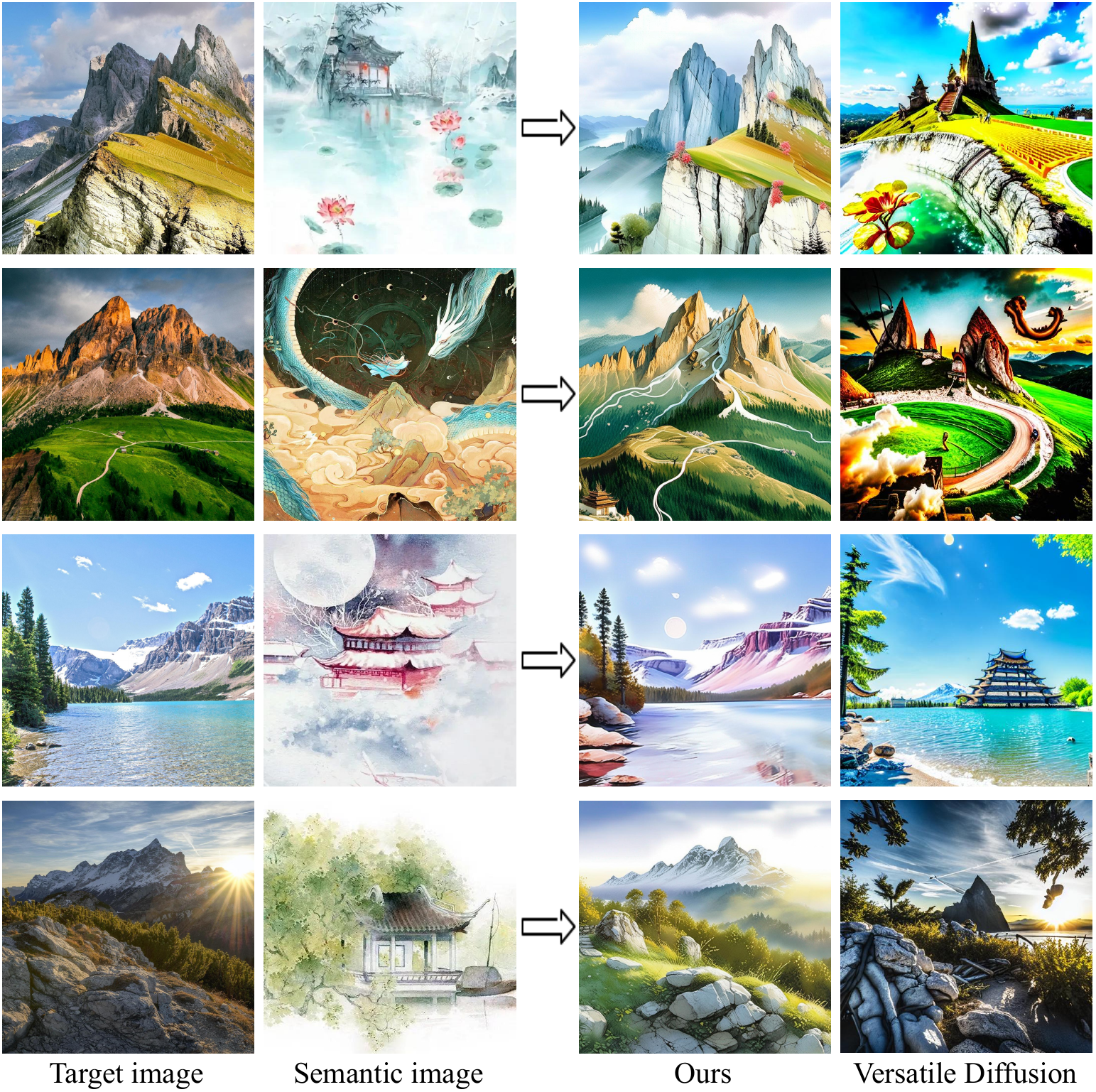}
    \caption{\textcolor{black}{Employ CreativeSynth to realize the presentation of the outputs of image style transfrer.}}
    \label{fig:style_transfer}
\end{figure}

\section{Discussion}
\subsection{Application}
\subsubsection{Image variation}
CreativeSynth's ability to distill stylistic and semantic nuances from a reference image allows it to create work that closely matches artistic qualities. Fig.~\ref{fig:variation} demonstrates the ability to generate similar results from a single reference image. 
As shown in Fig.~\ref{fig:variation}(c), we compared the results of CreativeSynth with the state-of-the-art multimodal guided generation method Versatile Diffusion~\cite{vd} that can lead to artifacts and distorted results, where the generated images may present inconsistencies in content.
These variations not only highlight CreativeSynth's adaptability in dealing with different artistic styles but also emphasize its utility in extending the user's creativity by offering the possibility of multiple styles from a single starting point.

\begin{figure}
    \centering
    \includegraphics[width=\linewidth]{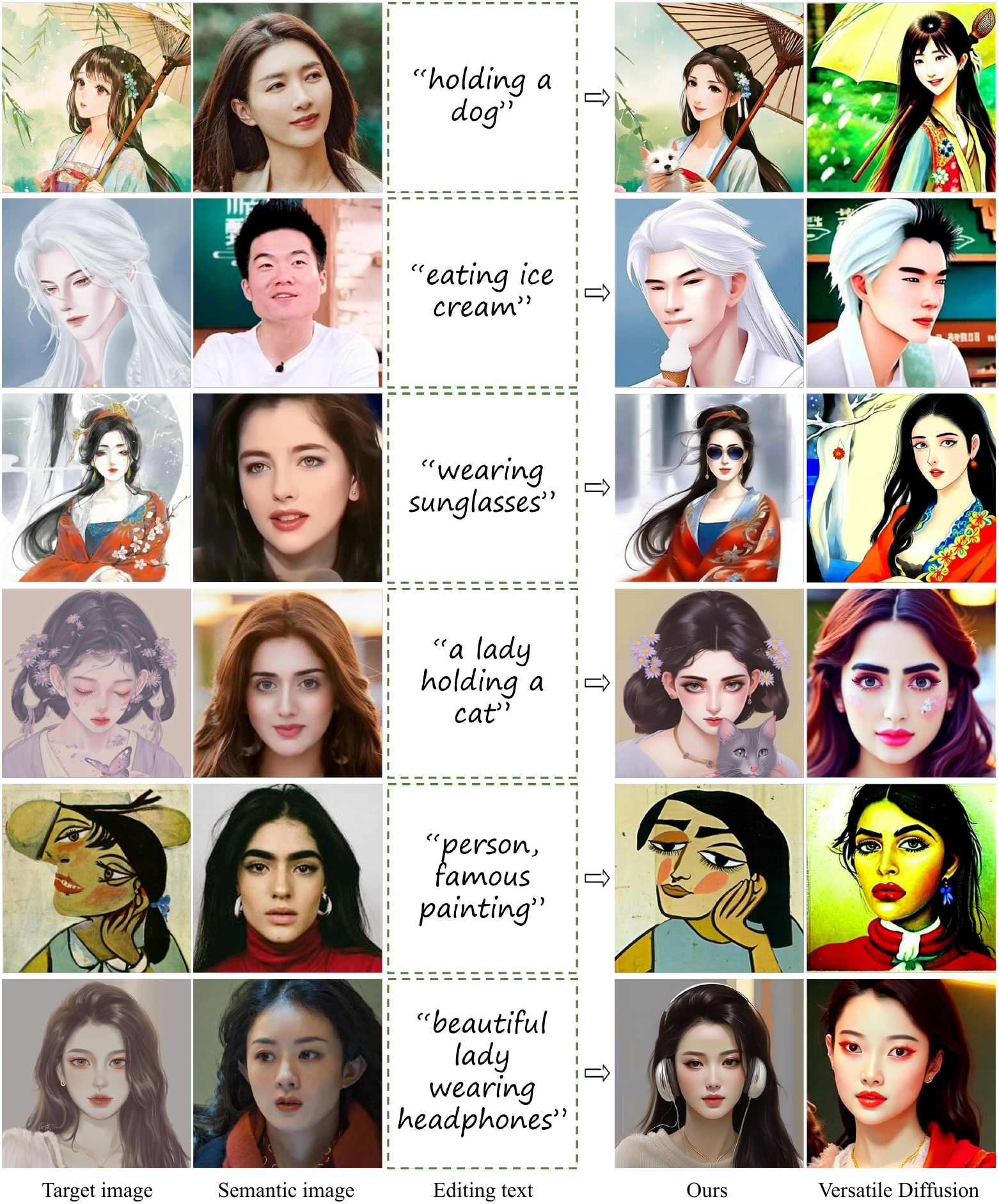}
    \caption{\textcolor{black}{Multimodal blending results. CreativeSynth can modify the content according to textual requirements based on the incorporation of semantic image information.}}
    \label{fig:2i+t}
\end{figure}

\subsubsection{Style transfer}
CreativeSynth specializes in fusion style and content for any given style of the target image, striking a balance between style transfer and preserving the original structure. 
While digital hand-drawn styles and anime provide a difficulty to standard style transfer methods, CreativeSynth may achieve style transfers, as seen in the example in Fig.~\ref{fig:style_transfer}. The remarkable capacity of CreativeSynth to produce digital art is demonstrated by its ability to imitate intricate artistic styles while preserving the recognizable elements of the original image.
In contrast, Fig.~\ref{fig:style_transfer}(d) results from Versatile Diffusion tend to introduce additional distorted content information, and they fail to maintain artistic effects well, often resulting in images with high saturation.

\begin{figure}
    \centering
    \includegraphics[width=\linewidth]{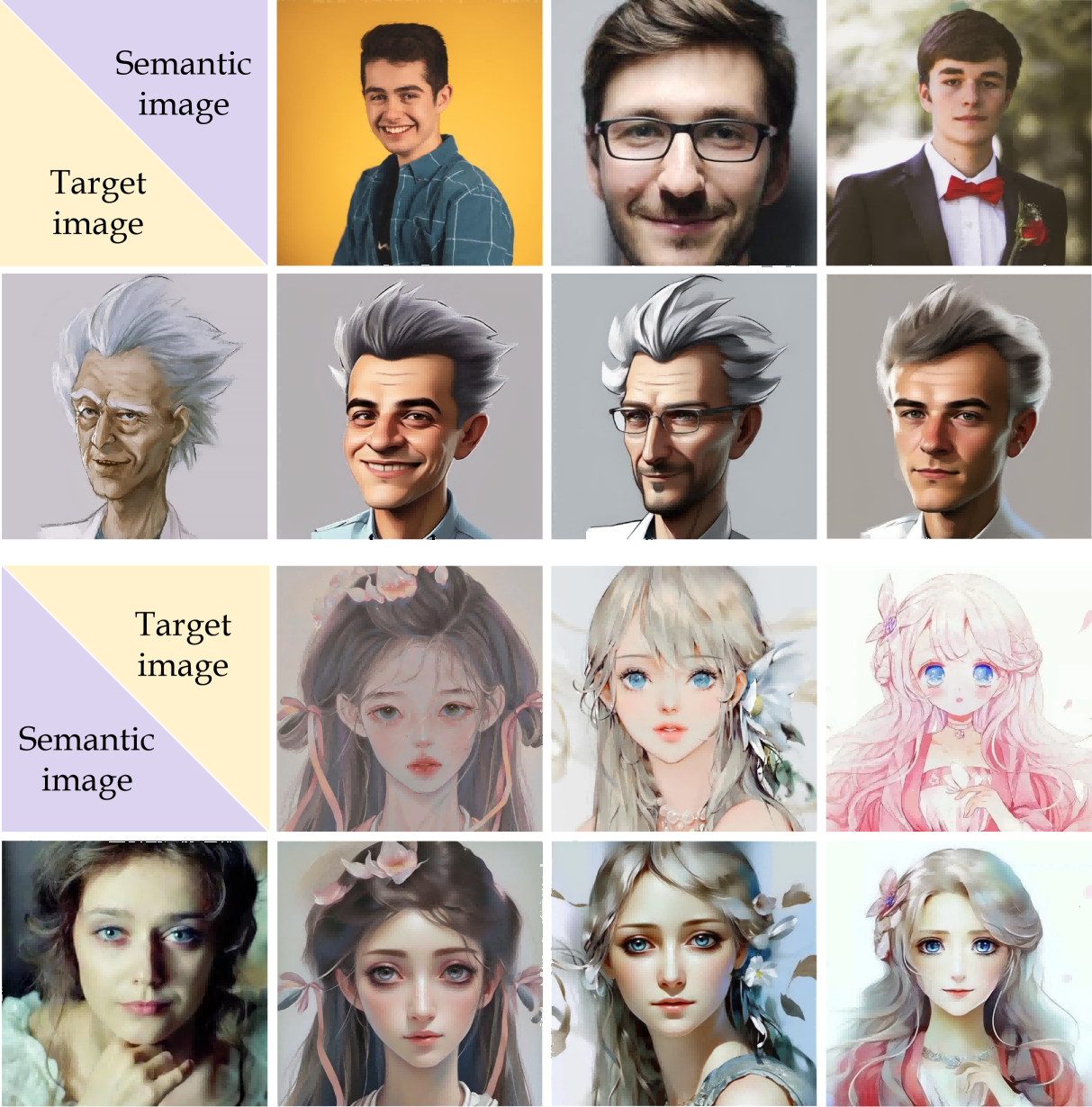}
    \caption{Diversity experiments results. The first two rows illustrate the results obtained by fixing the target image while replacing the semantic image, whereas the last two rows depict the results of fixing the semantic image and varying the target image. The results demonstrate the robustness and versatility of CreativeSynth.}
    \label{fig:fix_one}
\end{figure}

\subsubsection{multimodal fusion}
As shown in Fig.~\ref{fig:2i+t}, CreativeSynth extends to image generation and editing by integrating multimodal inputs.CreativeSynth demonstrates its innovative approach to art creation by fusing visual and textual prompts, thus introducing new semantic elements and transforming their content. 
Based on the user's creative thoughts, CreativeSynth may create concrete and semantically different artworks. The findings in Fig.~\ref{fig:2i+t} highlight the respective effects of semantic pictures and textual prompts on the created outcomes.
As demonstrated in Fig.~\ref{fig:2i+t}(d), CreativeSynth is able to maintain the artistic information better than the results of Versatile Diffusion in Fig.~\ref{fig:2i+t}(e), and generates more aesthetically pleasing and homogeneous results. Also, Versatile Diffusion could be improved in following the text editing information.
\begin{figure}
    \centering
    \includegraphics[width=0.8\linewidth]{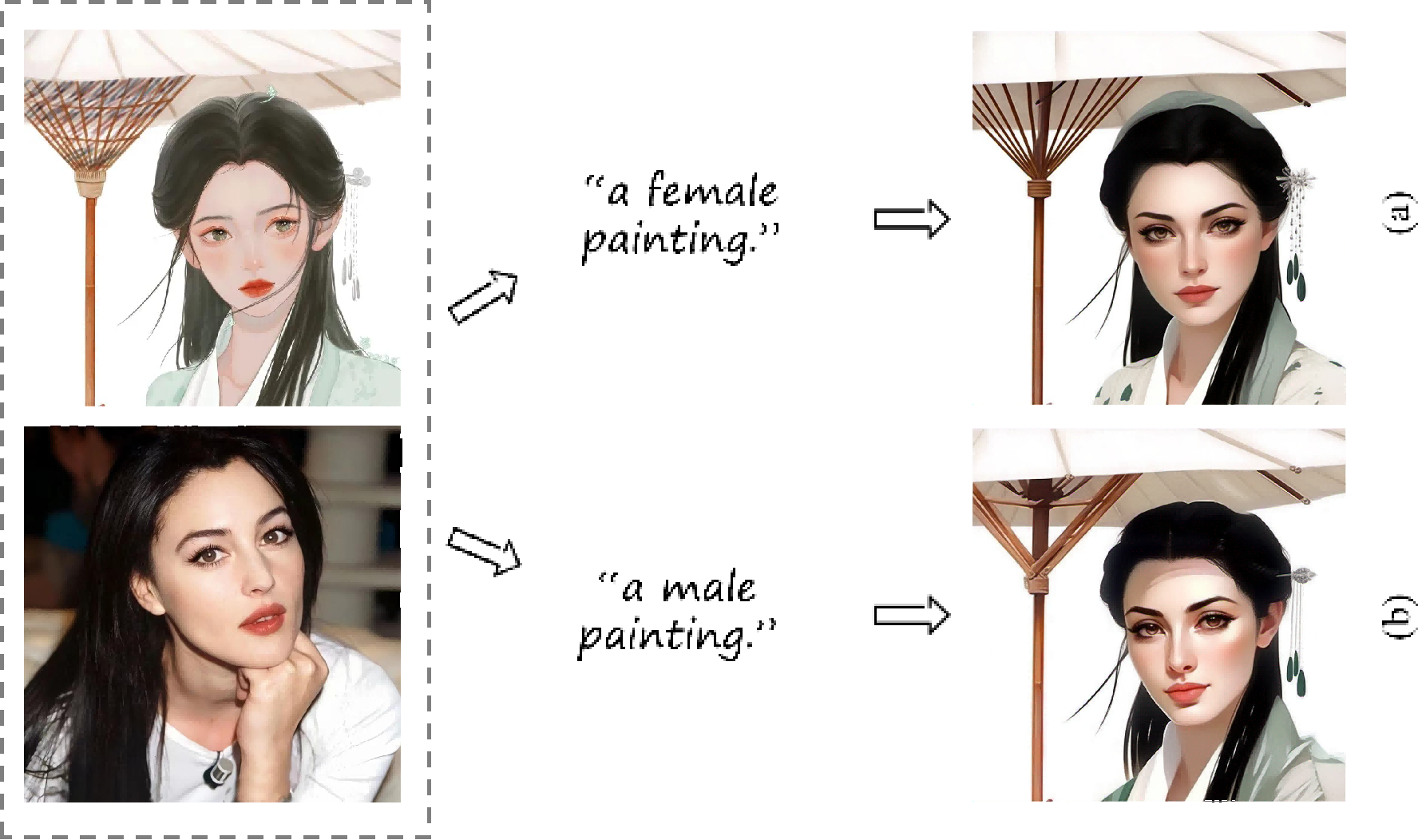}
    \caption{Results of the semantic discrepancy between semantic image and editing text. (a) Semantic consistency. (b) Semantic inconsistency.}
    \label{fig:semantic_conflict}
\end{figure}

\begin{figure}
    \centering
    \includegraphics[width=\linewidth]{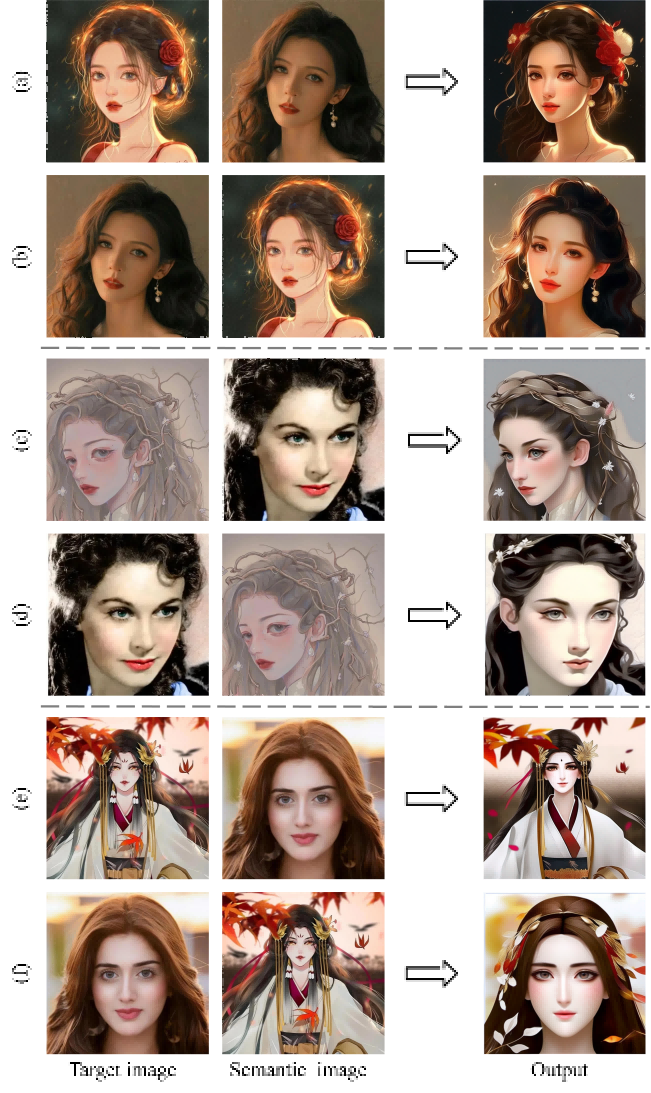}
    \caption{Experiments on exchanging input images between target image and semantic image. When the target image is an artistic image and the semantic image is a natural image (see (a), (c), and (e)), the process performs an image fusion task, such as embedding a human face into a painting. Conversely, when the target image is a natural image and the semantic image is an artistic image (see (b), (d), and (f)), the process executes a stylization task, introducing the style of the painting to the natural image.}
    \label{fig:change_place}
\end{figure}

\begin{figure*}
    \centering
    \includegraphics[width=0.9\linewidth]{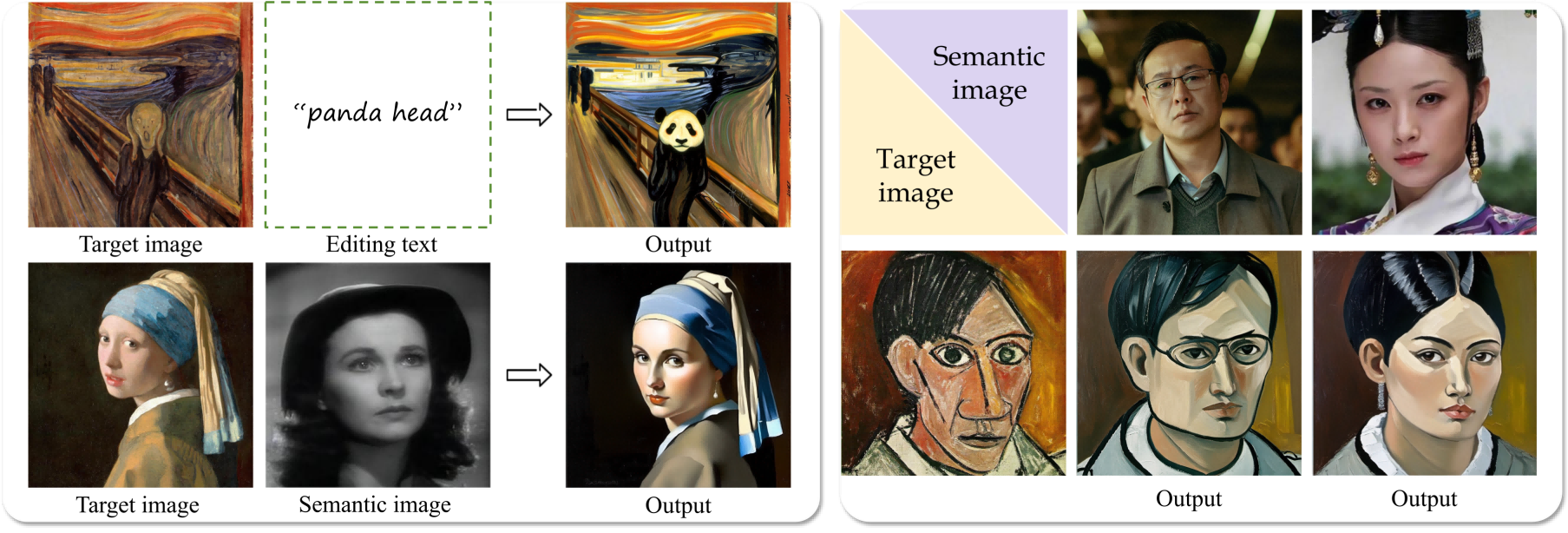}
    \caption{Bad cases. CreativeSynth is lacking in some of the special art textures, with a less prominent style that tends to produce smooth results.}
    \label{fig:badcase}
\end{figure*}

\subsubsection{Fixed input response}
To further illustrate the versatility of CreativeSynth, Fig.~\ref{fig:fix_one} shows CreativeSynth's response to the fixed target image and semantic image input. When the target image is held constant, CreativeSynth shows remarkable flexibility in adapting to a wide range of semantic inputs, being able to subtly incorporate its style and composition while faithfully retaining its inherent semantic qualities. This enables the creation of personalized artworks that resonate with the user's stylistic and thematic preferences. Conversely, if the input is a single semantic image, CreativeSynth can reinterpret the same subject in multiple styles, displaying a variety of features. The diversity and adaptability of the results shown in Fig.~\ref{fig:fix_one} highlight the versatility and robustness of CreativeSynth.

\subsection{Semantic conflict}
To deeply explore the interaction between semantic images and text prompts for their effects when they conflict, we conducted a series of experiments. As shown in Fig.~\ref{fig:semantic_conflict}, under the same image input condition, we can observe two different results. In Fig.~\ref{fig:semantic_conflict}(a), we input text prompts that are consistent with the semantic image, and the results show that the image generation is consistent with the expectation, indicating that the system can efficiently combine the two types of information to generate the target image when the textual and image information are coordinated.

Conversely, the scenario shifts in Fig.~\ref{fig:semantic_conflict}(b), where a text prompt discordant with the semantic image—specifically, ``male'' is deliberately introduced. This resulted in a notable phenomenon: despite the semantic image depicting a female, the textual information led to the enhancement of three-dimensionality in the female facial features, marked by increased shadows, grooves, and angles. This indicates that the effect of textual information on the image is significant, even if it conflicts with the semantic image.
It is worth noting that the final generated image retains the main female features despite the conflict between the text prompt and the semantic image, suggesting that the influence of the semantic image on the final result is still more significant in our CreativeSynth system. This finding suggests that when using such systems, parameter adjustments may be needed to properly balance the effects of textual and image information on the final generated results to achieve optimal image generation.

\subsection{Swapping Input Images}
To explore more possibilities of CreativeSynth inputs, we broke the inherent pairing of artistic image as target image and real image as semantic image. As shown in Fig.~\ref{fig:change_place}, we swapped the channels of the input image. From the results of rows 2, 4, and 6 in Fig.~\ref{fig:change_place}, the facial features of the people in the art image can be reproduced in the real image to a certain extent, but it is difficult to retain the authenticity of the faces, and the results are still more in favor of the art paintings.
CreativeSynth is primarily designed and optimized as an art generation model rather than a real face generation model. 
When it comes to handling the merger of realistic and creative pictures, CreativeSynth's particular powers and limits are demonstrated by this experimental method of channel-switching.
Although it performs well in mimicking features in artistic images, there are still some challenges in maintaining the point of face realism.
In our future research and applications, we can try to create artistic images without losing the sense of realism by using different combinations of inputs and parameter adjustments.

\subsection{Failure cases}
Although CreativeSynth is capable of working with a wide range of art styles, including Drawing and Cubism, it may not be accurate enough in capturing specific elements of certain styles, such as unusual shape variations and brushstroke characteristics, as shown in Fig.~\ref{fig:badcase}. This is demonstrated when attempting to transform or blend styles with complex textures and brushstroke expressions, such as Impressionism, the resulting artwork may be too smooth and lack the vigour and roughness of the brushstrokes typical of the style. This excessive smoothness and clarity may diminish the expressiveness of the work, especially when attempting to retain or emphasize specific stylistic features in the original artwork. 
To address these issues, future research may need to explore more refined style-capturing techniques, as well as improved algorithms that can better understand and reproduce detailed features, such as the variety of brushstrokes and the complexity of textures, in various art styles. By analyzing and modeling these details in greater depth, CreativeSynth, and similar tools may be able to support a wide range of artistic expression and style reproduction more effectively, providing artists and designers with a richer set of creative tools.

\section{Conclusions and Future Work}
In this paper, we present CreativeSynth, a unified framework designed to enable creative fusion and synthesis of visual artworks. 
\textcolor{black}{
The primary aim is to infuse multimodal semantic information into artworks using our novel Cross-Art-Attention mechanism without altering model parameters. This innovative approach ensures the preservation of the art pieces' inherent themes, emotions, and narratives, transcending a mere overlay of style onto natural images, by integrating semantic and aesthetic information through a unique semantic fusion mechanism and style alignment technique.
}
In this way, each synthesized work is not only a visual fusion, but also an intersection of meaning and story; with a strong personality, a unique visual narrative, and an exclusive emotional depth. Experimental results have shown that CreativeSynth is not only popular for its visual results, but also highly effective in executing user-specific artistic editorial intent. In the future, we plan to apply this approach to different image generation architectures and to broaden its application to encompass other forms of media, such as video. With subsequent refinements and applications, our approach will help creators realize creative expression like never before.

\bibliographystyle{IEEEtran}
\bibliography{references}

\newpage
\begin{IEEEbiography}[{\includegraphics[width=1in,height=1.25in,clip,keepaspectratio]{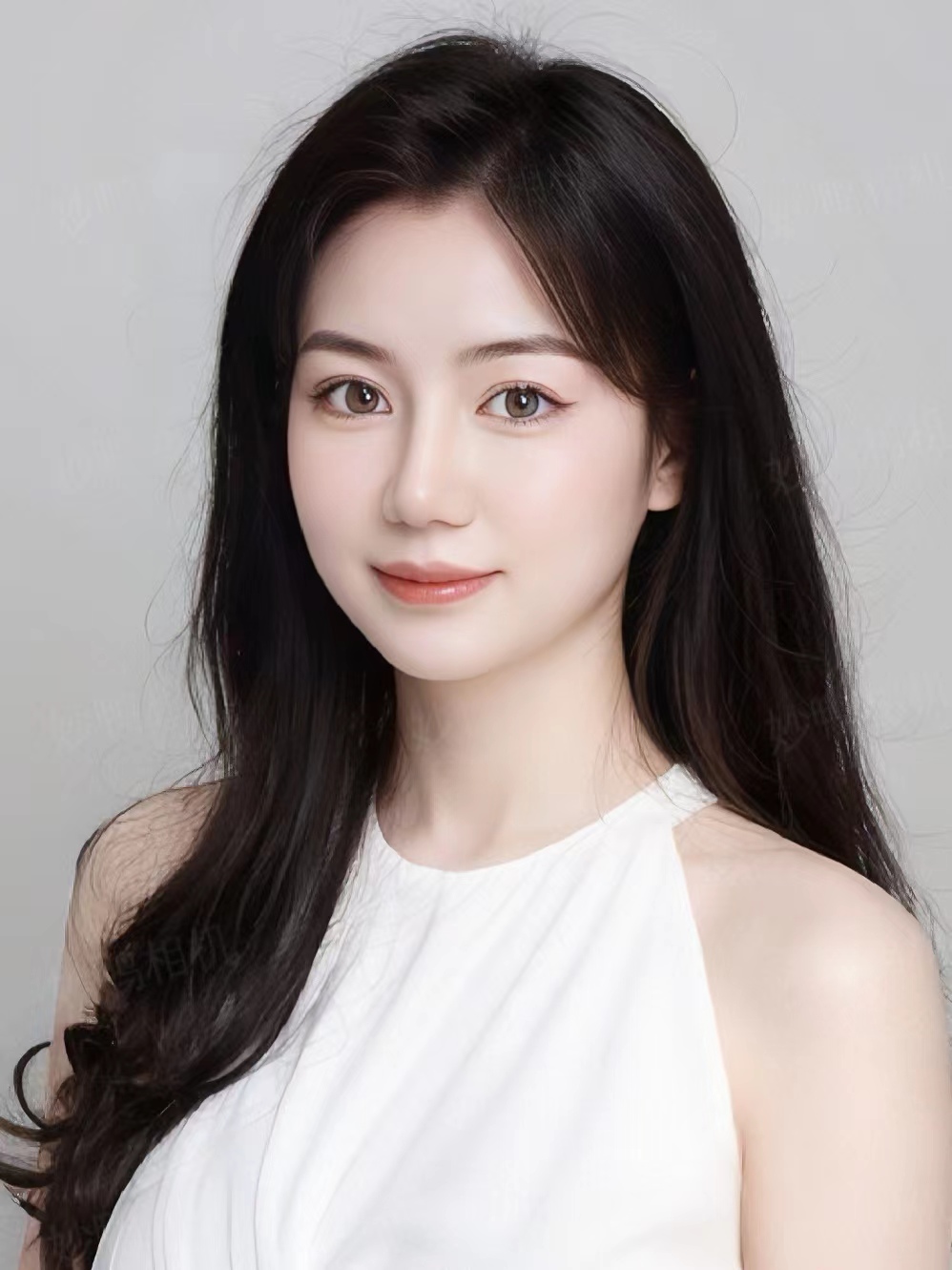}}]{Nisha Huang} received her B.S. degree in Aeronautical Information Engineering from Beijing University of Aeronautics and Astronautics in 2021 and her M.S. degree in Artificial Intelligence from the University of Chinese Academy of Sciences in 2024. Currently, she is pursuing her PhD in Electronic Information at Tsinghua University. And she is conducting research at the PengCheng Laboratory. Her research interests include generative models, multimedia analysis, and computer vision.
\end{IEEEbiography}
\begin{IEEEbiography}[{\includegraphics[width=1in,height=1.25in,clip,keepaspectratio]{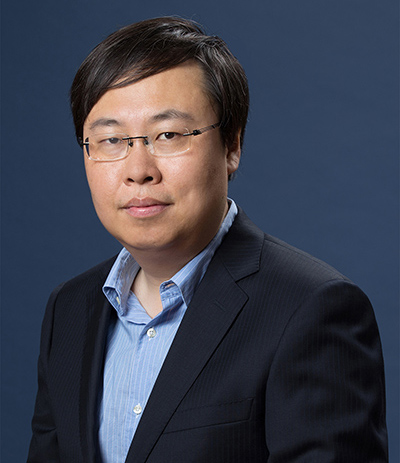}}]{Weiming Dong}
(Member, IEEE) is a Professor at the State Key Laboratory of Multimodal Artificial Intelligence Systems (MAIS), Institute of Automation, Chinese Academy of Sciences. He received his BSc and MSc degrees in 2001 and 2004, both from Tsinghua University, China. He received his PhD in Computer Science from the University of Lorraine, France, in 2007. His research interests include visual media generation, computational art, and generative artificial intelligence.
\end{IEEEbiography}
\begin{IEEEbiography}[{\includegraphics[width=1in,height=1.25in,clip,keepaspectratio]{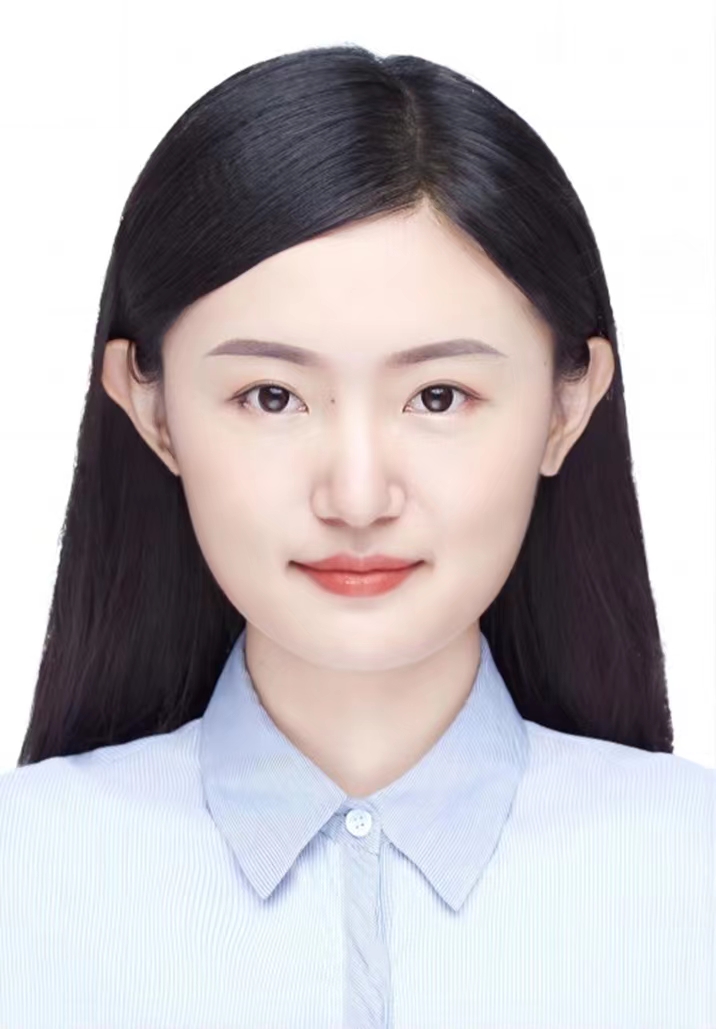}}]{Yuxin Zhang} received B.Sc. degree in Automation from Tsinghua University, Beijing, China, in 2020. She is now a Ph.D. candidate at the State Key Laboratory of Multimodal Artificial Intelligence Systems (MAIS), Institute of Automation, Chinese Academy of Sciences, and the School of Artificial Intelligence at the University of Chinese Academy of Sciences. Her research interests include computer vision, computer graphics, and machine learning.
\end{IEEEbiography}
\begin{IEEEbiography}[{\includegraphics[width=1in,height=1.24in,clip,keepaspectratio]{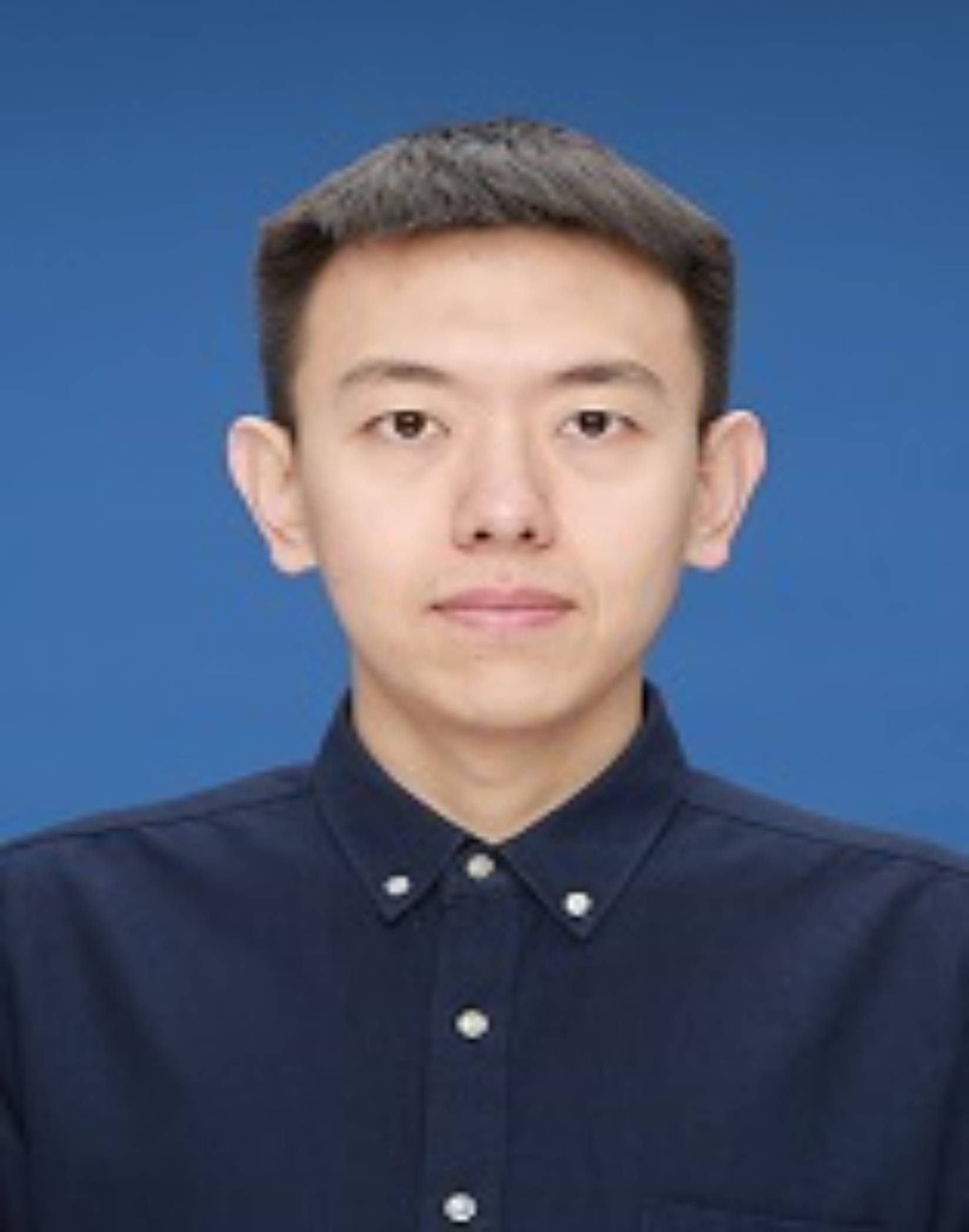}}]{Fan Tang} received the B.Sc. degree in computer science from North China Electric Power University, Beijing, China, in 2013, and the Ph.D. degree from Institute of Automation, Chinese Academy of Sciences, Beijing, in 2019. He is an Associate Professor at the School of Computer Science and Technology, University of Chinese Academy of Sciences. His research interests include computer graphics, computer vision, and machine learning.
\end{IEEEbiography}
\begin{IEEEbiography}[{\includegraphics[width=1in,height=1.24in,clip,keepaspectratio]{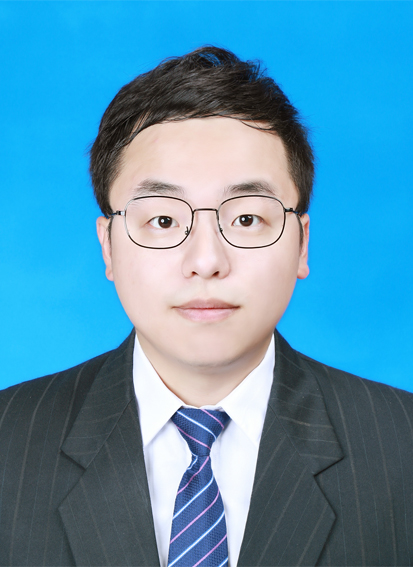}}]{Ronghui Li} received Master's degree in Control Engineering from Northeastern University, Shenyang, China in 2022. He is now a Ph.D. candidate at Shenzhen International Graduate School, Tsinghua University. His research interests include generative models, computer vision, and computer graphics.
\end{IEEEbiography}

\begin{IEEEbiography}[{\includegraphics[width=1in,height=1.25in,clip,keepaspectratio]{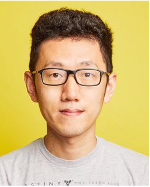}}]{Chongyang Ma}
received a B.S. degree from the Fundamental Science Class (Mathematics and Physics) of Tsinghua University in 2007 and a Ph.D. degree in Computer Science from the Institute for Advanced Study of Tsinghua University in 2012. He is currently a Tech Lead Manager at ByteDance. His research interests include computer graphics and computer vision.
\end{IEEEbiography}

\begin{IEEEbiography}[{\includegraphics[width=1in,height=1.25in,clip,keepaspectratio]{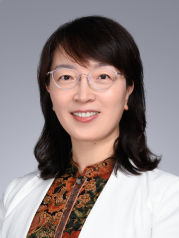}}]{Xiu Li}
(Member, IEEE) received a Ph.D. degree in computer-integrated manufacturing from the Nanjing University of Aeronautics and Astronautics, Nanjing, China, in 2000. From then to 2002, she served as a Post-Doctoral Fellow at the Department of Automation, Tsinghua University, Beijing, China. From 2003 to 2010, she served as an Associate Professor at the Department of Automation, at Tsinghua University. Since 2016, she has been a Full Professor at the Tsinghua Shenzhen International Graduate School, Tsinghua University, Shenzhen, China. Her research interests are in the areas of data mining, deep learning, computer vision, and image processing.
\end{IEEEbiography}

\begin{IEEEbiography}[{\includegraphics[width=1in,height=1.24in,clip,keepaspectratio]{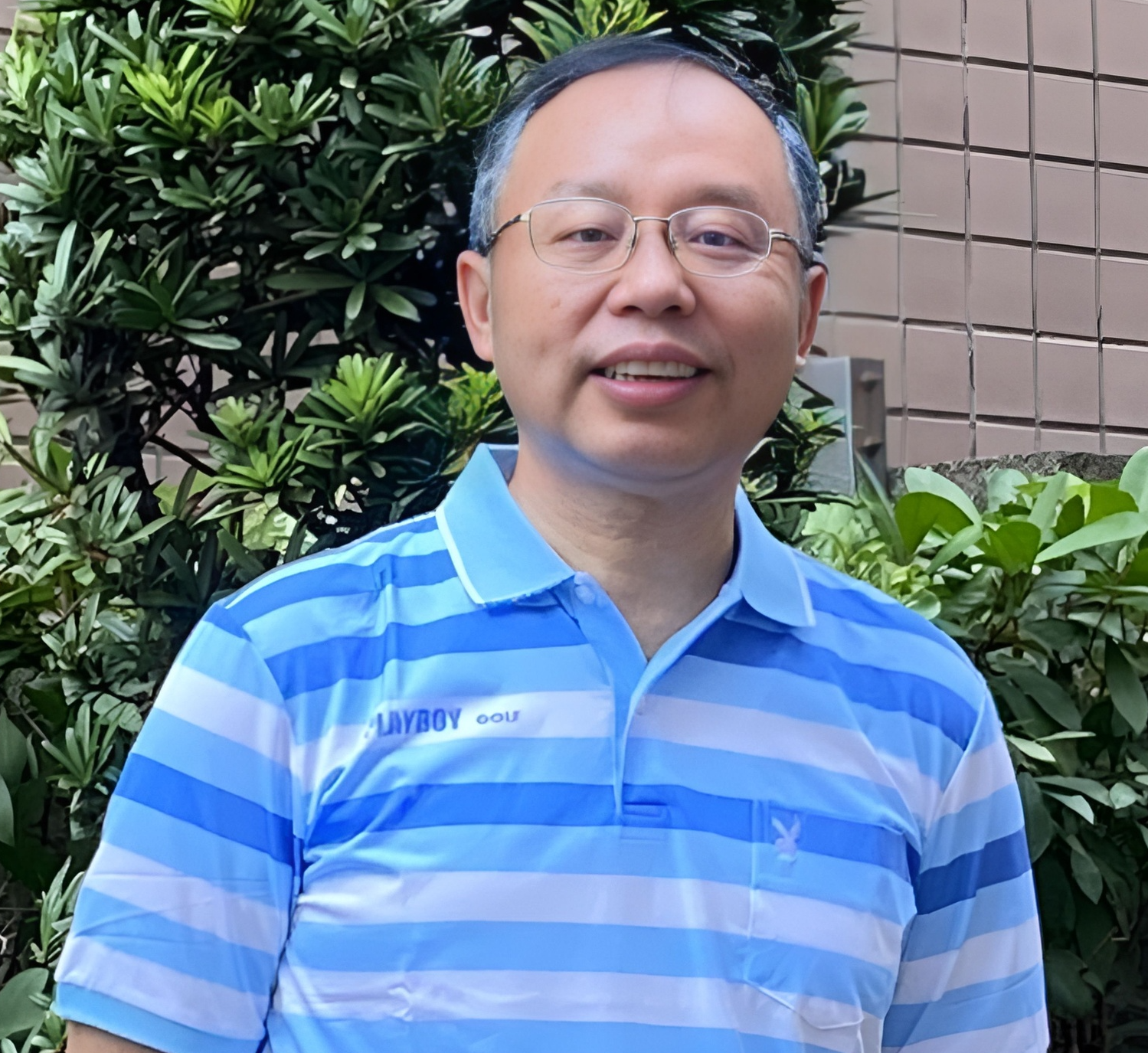}}]{Tong-Yee Lee} (Senior Member, IEEE) received the Ph.D. degree in computer engineering from Washington State University, Pullman, in May 1995. He is currently a Chair Professor in the Department of Computer Science and Information Engineering, at National Cheng-Kung University, Tainan City, Taiwan. He leads the Computer Graphics Laboratory, National Cheng-Kung University (http://graphics.csie.ncku.edu.tw). His current research interests include computer graphics, non-photorealistic rendering, medical visualization, virtual reality, and media resizing. He is a Senior Member of the IEEE and a Member of the ACM. He also serves on the editorial boards of the IEEE Transactions on Visualization and Computer Graphics.
\end{IEEEbiography}

\begin{IEEEbiography}[{\includegraphics[width=1in,height=1.25in,clip,keepaspectratio]{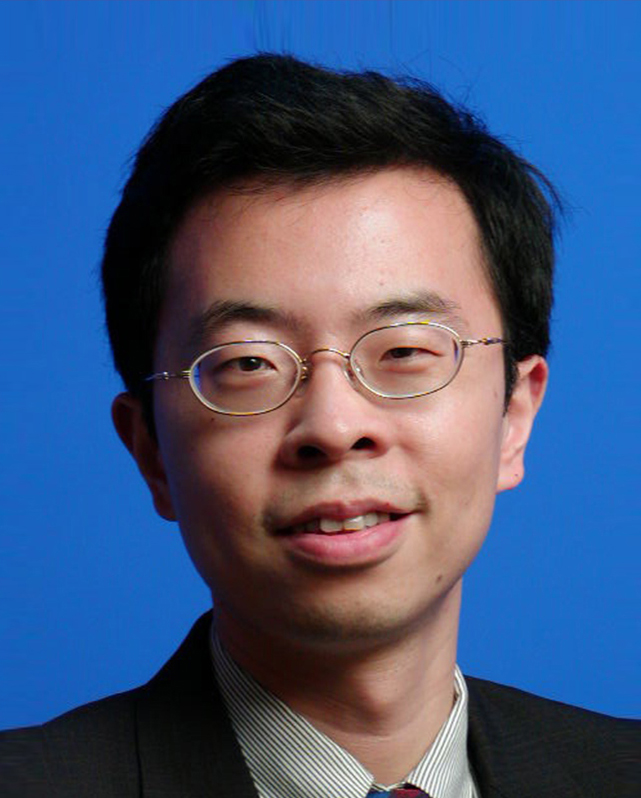}}]{Changsheng Xu}
(Fellow, IEEE) is a Professor at the State Key Laboratory of Multimodal Artificial Intelligence Systems (MAIS), Institute of Automation, Chinese Academy of Sciences.
Dr. Xu received the Best Associate Editor Award of ACM Transactions on Multimedia Computing, Communications, and Applications in 2012 and the Best Editorial Member Award of ACM/Springer Multimedia Systems Journal in 2008. He has served as an Associate Editor, a Guest Editor, a General Chair, a Program Chair, an Area/Track Chair, a Special Session Organizer, a Session Chair, and a Transactions on Professional Communication (TPC) Member for over 20 IEEE and ACM prestigious multimedia journals, conferences, and workshops. He is the International Association for Pattern Recognition (IAPR) Fellow and the ACM Distinguished Scientist.
\end{IEEEbiography}

\end{document}